\documentclass[10pt,twocolumn,letterpaper]{article}

\usepackage[pagenumbers]{cvpr}   

\usepackage{colortbl}
\usepackage{pifont}
\usepackage{multirow}
\usepackage[export]{adjustbox}
\usepackage{tikz}
\usetikzlibrary{positioning,arrows.meta}
\definecolor{rowgrey}{gray}{0.93}
\usepackage[most]{tcolorbox}
\definecolor{famMech}{HTML}{4C78C8}
\definecolor{famFluid}{HTML}{2E9FB0}
\definecolor{famSpat}{HTML}{E08A3C}
\definecolor{famOccl}{HTML}{8A63C9}
\definecolor{famTemp}{HTML}{D4A62A}
\definecolor{famAgent}{HTML}{3FA05A}
\definecolor{famRules}{HTML}{D0566C}
\definecolor{okgreen}{HTML}{2E8B4A}
\definecolor{badred}{HTML}{C0392B}
\newtcolorbox{famcard}[2][]{enhanced, equal height group=fam, colback=white,
  colframe=#2!65!white, colbacktitle=#2!18!white, coltitle=black, boxrule=0.6pt,
  arc=2.5pt, left=3pt, right=3pt, top=3pt, bottom=2pt, toptitle=1.5pt,
  bottomtitle=1.5pt, halign title=center, fonttitle=\footnotesize\bfseries, #1}
\newcommand{\stripframe}[1]{\makebox[0.316\linewidth]{\includegraphics[width=0.316\linewidth,height=0.1778\linewidth,keepaspectratio]{#1}}}
\newcommand{\filmstrip}[1]{{\centering\setlength{\fboxsep}{1.2pt}%
  \colorbox{black!80}{\stripframe{figures/frames/#1_0.jpg}\hspace{1pt}%
  \stripframe{figures/frames/#1_1.jpg}\hspace{1pt}%
  \stripframe{figures/frames/#1_2.jpg}}\par}\vspace{3pt}}

\newcommand{\cmark}{\ding{51}}
\newcommand{\xmark}{\ding{55}}
\newcommand{\pmark}{(\ding{51})}

\newcommand{\bench}{GRIP-V\xspace}



\definecolor{cvprblue}{rgb}{0.21,0.49,0.74}
\usepackage[pagebackref,breaklinks,colorlinks,allcolors=cvprblue]{hyperref}

\def\confName{CVPR}
\def\confYear{2027}

\title{Watch, Don't Glance: A Procedurally Generated Video Benchmark\\
for Temporal Geometric and Physical Reasoning}

\author{%
Hamza Hanif$^{1}$\quad
Muhammad Bilal Khan$^{2}$\quad
Inam ur Rehman$^{3}$\quad
Amgad Muneer$^{4}$\quad
Bilal Jawaid$^{5}$\\
Muhammad Maaz Uddin$^{6}$\quad
Anas Zafar$^{4}$\quad
Shahzaib Abbas$^{7}$\quad
Ahmad Abdullah Qamar$^{8}$\\
Aizan Zafar$^{5}$\quad
Aman Chadha$^{9}$\quad
Hong Yan$^{10}$\quad
Rizwan Qureshi$^{11}$\\[6pt]
{\small
$^{1}$Western University\quad
$^{2}$University of Delaware\quad
$^{3}$Habib University}\\
{\small
$^{4}$The University of Texas MD Anderson Cancer Center\quad
$^{5}$Independent\quad
$^{6}$Salim Habib University}\\
{\small
$^{7}$Technische Universit\"at Dortmund\quad
$^{8}$Lahore Grammar School Islamabad\quad
$^{9}$Google}\\
{\small
$^{10}$City University of Hong Kong\quad
$^{11}$Massachusetts General Hospital}\\[6pt]
{\small
\href{https://grip-v.github.io/}{\texttt{grip-v.github.io}}\quad
Code: \href{https://github.com/grip-v/grip-v-scripts}{\texttt{github.com/grip-v/grip-v-scripts}}}\\
{\small Data: \href{https://huggingface.co/datasets/hamzahanif/GRIP-V}{\texttt{huggingface.co/datasets/hamzahanif/GRIP-V}}}
}

\begin{document}
\maketitle
\begin{abstract}
Robots and other physical-AI systems increasingly rely on vision-language
models (VLMs) to perceive a scene and predict how it will evolve. Yet many
questions in video benchmarks can be answered from a single frame, so a high
score does not show that a model followed the scene over time. We introduce
\textbf{\bench}, the video version of GRIP~\cite{jawaid2026grip}: 40
procedurally generated sub-benchmarks in seven reasoning families, from
mechanics and fluids to occlusion, agents, and rule tracking, with
$2{,}500$ clips per sub-benchmark and ten questions per clip ($1$ million
questions). Every answer is computed from the simulated scene state and
checked by a validator. Questions climb a five-rung ladder from observation
through relations, measurement, and multi-step integration to
counterfactual reasoning, and 81.1\% are flagged as requiring motion
because they ask about the trajectory, such as event counts or order,
rather than the end state. We evaluated eleven open-weight VLMs on a
$100{,}000$-question subset. The best model answers 55.6\% of the
questions, against a constant-answer baseline of 28.1\%. Shown only the
final frame, three of five tested models lose 14--15 points of
baseline-adjusted accuracy, so they do use motion. Accuracy, however,
falls by up to 37 points from the first rung to the last, eight of eleven
models are within 6 points of the baseline on counterfactuals, and no
model reliably detects injected rule violations. Current VLMs can follow
how a scene unfolds, but not yet predict how it would change under an
intervention or notice when it breaks the rules, the two abilities a robot
needs to plan and to monitor its own actions.
\end{abstract}
\section{Introduction}
\label{sec:intro}

\begin{figure}[t]
\centering
\resizebox{\linewidth}{!}{\input{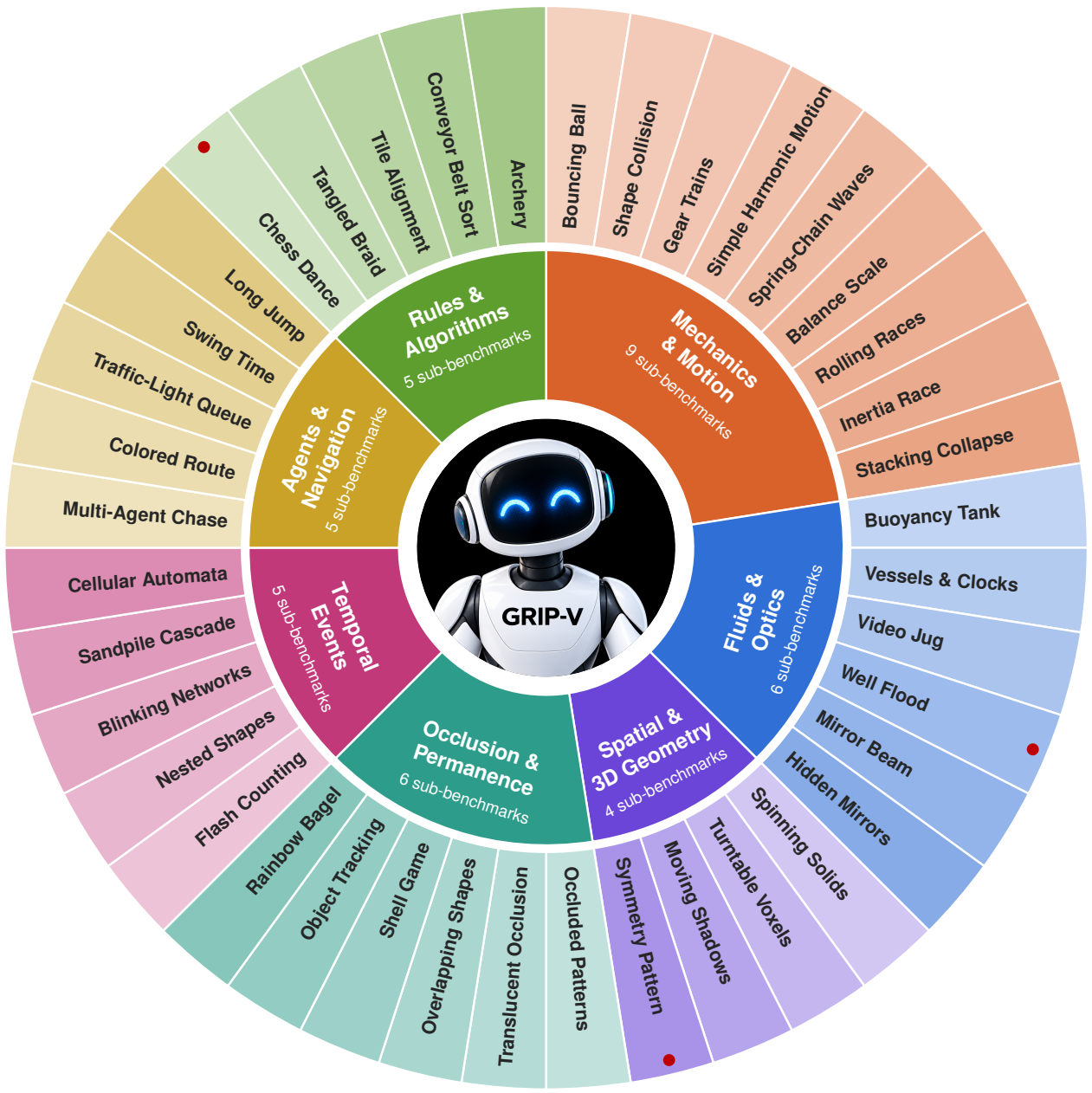}}
\caption{\textbf{The \bench taxonomy.} 40 video sub-benchmarks (outer ring)
grouped into seven reasoning families (inner ring), each with $2{,}500$
procedurally generated clips and ten questions per clip. A red dot marks the 3
sub-benchmarks with injected single-violation clips.}
\label{fig:wheel}
\end{figure}

Vision-language models (VLMs) increasingly serve as the perception and
reasoning front end of robots and other physical-AI systems. Embodied language
models and vision-language-action policies build robot control on VLM
backbones~\cite{driess2023palme,brohan2023rt2,kim2024openvla}, and world
foundation models learn to predict how scenes evolve~\cite{nvidia2025cosmos}.
Such systems must follow objects that move, collide, pour, and disappear
behind occluders, and predict what an action would change. A robot that pours
must know how much is left in the jug~\cite{wang2026liquidlab}; one that stacks
blocks must know when the tower will tip.

Testing these abilities requires video, but video alone is not enough. Many
questions in video benchmarks can be answered from one frame or from a
language prior, so a high score does not show that a model tracked
anything~\cite{liu2024tempcompass,mangalam2023egoschema}. GRIP~\cite{jawaid2026grip}
addressed a similar problem for images with 34 procedurally generated domains,
exact ground truth, and a five-rung question ladder. \bench brings this method
to video and asks whether a model can follow a scene as it changes.

The design rule of \bench is that, beyond direct observation, a question
should ask about something no single frame shows (\cref{fig:teaser}): where a
hidden cup went, which strand crossed which in a braid, the rotation direction
of a gear train, how often an object flashed while hidden, the
round in which a chess ``dance'' broke the rules, or which ball swallowed which
in a chase. Each question carries a motion flag; it is set
for 81.1\% of questions and for 87--95\% on rungs L2--L5. The unflagged
questions are mostly L1 questions that check whether a model recognizes the
scene at all. Every clip is drawn frame by frame from an analytical model or
a deterministic simulation, and every answer is computed from the same scene
state, not read from pixels. A validator for each sub-benchmark re-derives
every answer from the drawn scene and checks decoded frames against the
expected state; it finds zero mismatches. Our contributions are:

\begin{itemize}
\setlength{\itemsep}{1pt}
\item \textbf{\bench}: 40 video sub-benchmarks in seven reasoning families
(\cref{fig:wheel}), each with $2{,}500$ clips and ten questions per clip
($1{,}000{,}000$ questions), with 185 scenario variants such as brakes, leaks,
reversals, and hidden rules (\cref{sec:variants}).
\item \textbf{A shared five-rung ladder}: observation, relation, measurement,
multi-step integration, and counterfactual reasoning, applied to every
sub-benchmark (\cref{sec:ladder}).
\item \textbf{Injected violations}: in three sub-benchmarks, 20--45\% of clips
contain exactly one illegal event (a chess move, a symmetry break, or an
impossible flow), and the model must say whether and where it happened
(\cref{sec:violations}).
\item \textbf{Motion controls}: the same questions asked with only the final
frame, or with shuffled frames, measure how much a model gains from motion
and from frame order (\cref{sec:setup}).
\item \textbf{An evaluation} of eleven open-weight VLMs from seven families
(7B to 180B parameters) on a fixed $100{,}000$-question subset, with a human
baseline on $7{,}720$ questions (\cref{sec:results}).
\end{itemize}

\noindent
The best model, Qwen3.6-27B, answers 55.6\% of the questions; the weakest
answers 31.2\%, against a constant-answer baseline of 28.1\%. On
counterfactual questions, eight of eleven models are within 6 points of the
baseline after adjustment. Given only the final frame, three of five tested
models lose 14--15 points of adjusted accuracy, and shuffling the frames costs
the two strongest tested models about 10 points, so these models do use
motion. No model reliably detects the injected violations: models either
almost always answer that nothing went wrong or raise many false alarms.

\section{Related Work}
\label{sec:related}

\paragraph{Video understanding benchmarks.}
General-purpose video understanding suites such as MVBench~\cite{li2024mvbench},
Video-MME~\cite{fu2025videomme}, Perception Test~\cite{patraucean2023perception},
and EgoSchema~\cite{mangalam2023egoschema} cover a broad range of real-world
footage, but their labels come from human annotation or are adapted from
existing datasets, and many items can be answered from a few frames.
TempCompass~\cite{liu2024tempcompass} addresses this with conflicting-video
controls for speed and direction, and TOMATO~\cite{shangguan2025tomato}
reports how much models gain from many frames over one and from frame order.
VideoReasonBench~\cite{liu2026videoreasonbench} generates most of its clips
by program and asks models to recall, infer, and predict a hidden state.
Two concurrent benchmarks also target state tracking: VET-Bench~\cite{liu2026vetbench}
uses synthetic shell games with identical objects, on which VLMs score near
chance, and VSTAT~\cite{yu2026vstat} asks questions about synthetic and
real clips that no single frame answers.
\bench generates every clip, so the temporal dependency holds by
construction, and it covers 40 physical and rule-based domains under one
five-rung ladder, with final-frame and shuffled-frame controls.

\paragraph{Synthetic and physical video reasoning.}
CLEVRER~\cite{yi2020clevrer} and CATER~\cite{girdhar2020cater} demonstrated
the value of synthetic videos for causal and compositional temporal reasoning.
IntPhys~\cite{riochet2021intphys} and Physion~\cite{bear2021physion}
probe intuitive physics through possible/impossible events and outcome
prediction, building on earlier block-tower studies~\cite{lerer2016blocktowers}.
More recent VLM-oriented physics benchmarks~\cite{chow2025physbench,schulzebuschoff2026interaction}
evaluate physical-world understanding at scale. \bench combines three
elements: (i) broad coverage of rule-governed and symbolic dynamics,
including chess, rope braids, cellular automata, and flooding, alongside
Newtonian physics; (ii) state-derived ground truth with independent
answer validation and rendered-frame consistency checks; and
(iii) a shared five-level reasoning ladder across all 40 sub-benchmarks.

\paragraph{VLMs for robotics and physical AI.}
Embodied multimodal models such as PaLM-E~\cite{driess2023palme} and
vision-language-action models such as RT-2~\cite{brohan2023rt2} and
OpenVLA~\cite{kim2024openvla} build on pretrained language and vision
representations to support robotic perception, reasoning, and action.
World foundation models for physical AI~\cite{nvidia2025cosmos} learn
to predict or generate how scenes evolve. Evaluating these systems
through robot task success or generated video quality does not
directly isolate their reasoning capabilities. \bench provides a
complementary evaluation of temporal and physical reasoning through
questions about remaining liquid, gear reversals, and occluded
objects, scored against state-derived ground truth without a robot
in the loop. Recent robot-learning work points to the same gap: policies
conditioned on a single observation cannot resolve motion or task
phase~\cite{feng2026tempo}, lose task-relevant cues once they are
occluded~\cite{yang2026eventvla}, and depend on temporal rather than
single-frame structure in their visual
representations~\cite{yeom2026latents}, while VLM benchmarks for manipulation
and collision safety report partial physical reasoning but unreliable
judgments~\cite{maurya2026imbench,wang2026collision}. The Molmo2 family that
we evaluate also provides the embodied-reasoning backbone of the MolmoAct2
action model~\cite{fang2026molmoact2}.

\paragraph{Procedural reasoning and grounded evaluation.}
Procedurally generated benchmarks such as RAVEN~\cite{zhang2019raven} and
PGM~\cite{barrett2018measuring} demonstrate the value of controlled
structure and exact answers for abstract reasoning. GeoQA~\cite{chen2021geoqa},
MathVista~\cite{lu2024mathvista}, and MathVerse~\cite{zhang2024mathverse}
evaluate geometric and mathematical reasoning from static visual inputs.
VLMs also struggle with basic visual primitives~\cite{rahmanzadehgervi2024blind}
and can change correct answers under social pressure~\cite{sharma2024sycophancy,rahman2025pendulum}.
\bench extends the image-based GRIP benchmark~\cite{jawaid2026grip},
preserving its exact-ground-truth design and five-rung reasoning ladder
while introducing dynamic scenes and trajectory-dependent questions.

\cref{tab:comparison} summarizes the broader benchmark differences.

\begin{table}[t]
\caption{\textbf{Comparison with video reasoning benchmarks.}
\emph{Derived}: ground truth computed from the generating program.
\emph{Proc.}: procedurally generated. \emph{Ladder}: shared compositional
difficulty levels. \emph{1-frame}: explicit single-frame control.
\emph{Viol.}: localized rule-violation detection. \pmark: partial; some clips recorded rather than generated, fixed reasoning types or tasks rather than a graded ladder, a single-frame baseline for only one task or model, or clip-level detection without localization.}
\label{tab:comparison}
\centering
\footnotesize
\setlength{\tabcolsep}{3.2pt}
\begin{tabular}{@{}lccccc@{}}
\toprule
\textbf{Benchmark} & \textbf{Derived} & \textbf{Proc.} & \textbf{Ladder} & \textbf{1-frame} & \textbf{Viol.}\\
\midrule
\rowcolor{rowgrey}
MVBench~\cite{li2024mvbench}                   & \xmark & \xmark & \xmark & \xmark & \xmark\\
Video-MME~\cite{fu2025videomme}                & \xmark & \xmark & \xmark & \xmark & \xmark\\
\rowcolor{rowgrey}
Perception Test~\cite{patraucean2023perception}& \xmark & \xmark & \pmark & \pmark & \xmark\\
TempCompass~\cite{liu2024tempcompass}          & \xmark & \xmark & \xmark & \cmark & \xmark\\
\rowcolor{rowgrey}
TOMATO~\cite{shangguan2025tomato}              & \xmark & \xmark & \xmark & \cmark & \xmark\\
VideoReasonBench~\cite{liu2026videoreasonbench} & \cmark & \pmark & \cmark & \cmark & \xmark\\
\rowcolor{rowgrey}
CLEVRER~\cite{yi2020clevrer}                   & \cmark & \cmark & \pmark & \xmark & \xmark\\
CATER~\cite{girdhar2020cater}                  & \cmark & \cmark & \pmark & \pmark & \xmark\\
\rowcolor{rowgrey}
IntPhys~\cite{riochet2021intphys}              & \cmark & \cmark & \xmark & \xmark & \pmark\\
Physion~\cite{bear2021physion}                 & \cmark & \cmark & \xmark & \xmark & \xmark\\
\midrule
\rowcolor{rowgrey}
\textbf{\bench (ours)}                         & \cmark & \cmark & \cmark & \cmark & \cmark\\
\bottomrule
\end{tabular}
\end{table}

\section{The \bench Benchmark}
\label{sec:benchmark}

\subsection{Design principles}
\label{sec:principles}

\begin{figure*}[t]
\centering
\begin{minipage}[t]{0.242\textwidth}
\begin{famcard}[title={Mechanics \& motion}]{famMech}
\filmstrip{stacking_collapse}
{\scriptsize\textit{stacking collapse}\par\vspace{1pt}
\textbf{Q (L2).} Which way does the plank tip?\par\vspace{1pt}
\textbf{GT.} \texttt{to the left}\par}
\end{famcard}
\end{minipage}\hfill
\begin{minipage}[t]{0.242\textwidth}
\begin{famcard}[title={Fluids \& optics}]{famFluid}
\filmstrip{glasses_fill}
{\scriptsize\textit{glasses fill}\par\vspace{1pt}
\textbf{Q (L2).} Which jug tilts over a glass next, right after the zigzag jug tilts for the first time?\par\vspace{1pt}
\textbf{GT.} \texttt{the wavy jug}\par}
\end{famcard}
\end{minipage}\hfill
\begin{minipage}[t]{0.242\textwidth}
\begin{famcard}[title={Spatial \& 3D geometry}]{famSpat}
\filmstrip{spinning_polyhedra}
{\scriptsize\textit{spinning polyhedra}\par\vspace{1pt}
\textbf{Q (L2).} How does the turning speed of the solid on the right change during the clip?\par\vspace{1pt}
\textbf{GT.} \texttt{steady speed}\par}
\end{famcard}
\end{minipage}\hfill
\begin{minipage}[t]{0.242\textwidth}
\begin{famcard}[title={Occlusion \& permanence}]{famOccl}
\filmstrip{shell_game_tracking}
{\scriptsize\textit{shell game tracking}\par\vspace{1pt}
\textbf{Q (L2).} The first time the cup hiding the orange triangle moves, to which position does it go?\par\vspace{1pt}
\textbf{GT.} \texttt{bottom-middle}\par}
\end{famcard}
\end{minipage}\\[4pt]
\hspace{0.064\textwidth}\begin{minipage}[t]{0.242\textwidth}
\begin{famcard}[title={Temporal events}]{famTemp}
\filmstrip{blinking_networks}
{\scriptsize\textit{blinking networks}\par\vspace{1pt}
\textbf{Q (L2).} Of the nodes directly linked to the red star, which one lights up first?\par\vspace{1pt}
\textbf{GT.} \texttt{blue star}\par}
\end{famcard}
\end{minipage}\hfill
\begin{minipage}[t]{0.242\textwidth}
\begin{famcard}[title={Agents \& navigation}]{famAgent}
\filmstrip{multi_agent_chase}
{\scriptsize\textit{multi-agent chase}\par\vspace{1pt}
\textbf{Q (L2).} Which ball is the first to be caught?\par\vspace{1pt}
\textbf{GT.} \texttt{purple checkered}\par}
\end{famcard}
\end{minipage}\hfill
\begin{minipage}[t]{0.242\textwidth}
\begin{famcard}[title={Rules \& algorithms}]{famRules}
\filmstrip{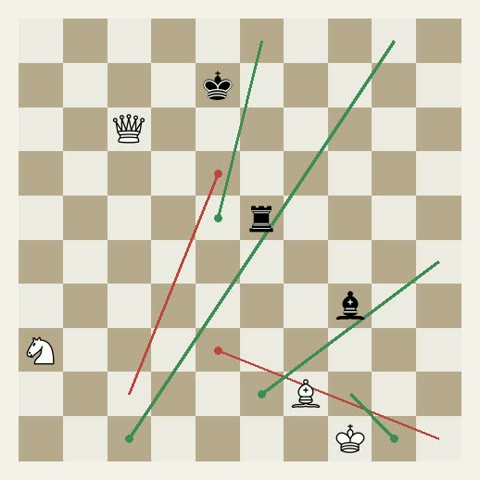}
{\scriptsize\textit{chess dance trace}\par\vspace{1pt}
\textbf{Q (L2).} Which piece captures the teal bishop (lands on its square, making it disappear)?\par\vspace{1pt}
\textbf{GT.} \texttt{the lime knight}\par}
\end{famcard}
\end{minipage}\hspace{0.064\textwidth}
\caption{\textbf{One \bench clip per reasoning family} (three frames of the first clip, left to right in time) with one of its ten questions (GT: ground truth; each
question has two to four options). Every answer depends on what happens
during the clip: the order of events, a direction of motion, or where a hidden
object went.}
\label{fig:teaser}
\end{figure*}

\paragraph{State-derived ground truth.}
Each clip is fully determined by a seeded parameter record. Its dynamics
come from closed-form expressions or deterministic simulations, for example
closed-form bounce heights for restitution, angular velocity propagated
through the gear graph, an event queue for flooding, and the rolling law
$a = g\sin\theta/(1+k)$ for the inertia race. The renderer draws every frame
from this scene state and encodes it as H.264 video
($1280\times720$, 24\,fps, at most 10\,s), and the answers are computed from
the same state. On-screen text is limited to a clock and, where needed,
ruler or grid labels, so no caption or counter gives the answer away.

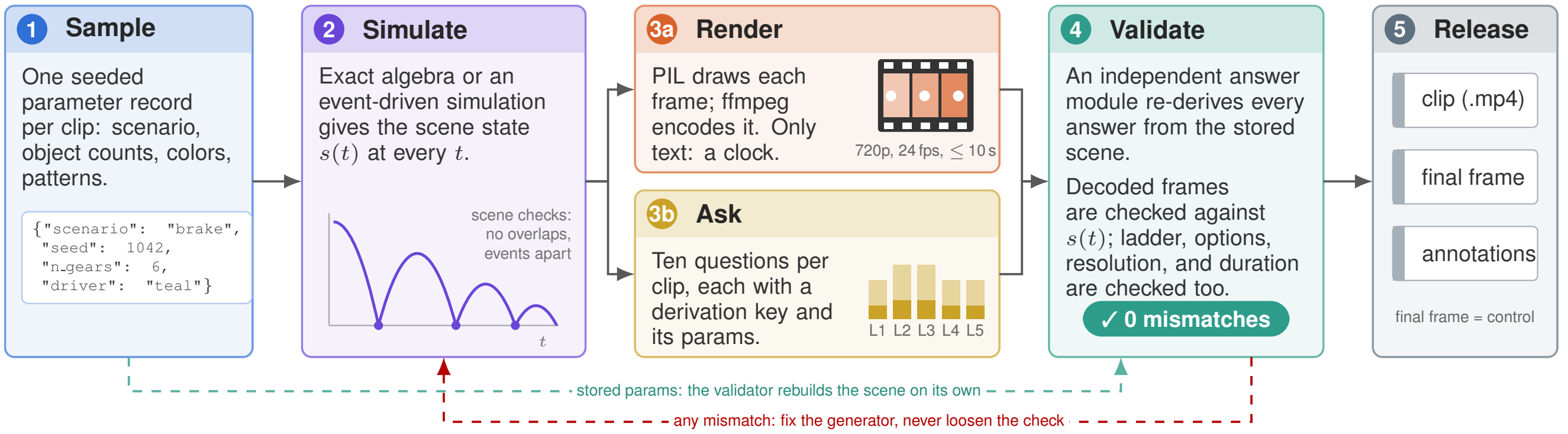
\begin{figure*}[t]
\centering
\resizebox{\linewidth}{!}{
\begin{tikzpicture}[
  font=\sffamily,
  flow/.style={-{Latex[length=2mm,width=1.6mm]}, black!65, line width=0.7pt},
  body/.style={anchor=north west, align=left, inner sep=0pt, font=\sffamily\scriptsize,
               text=black!80, execute at begin node={\hyphenpenalty=10000\exhyphenpenalty=10000}},
]
\definecolor{pSeed}{HTML}{2F6FD6}
\definecolor{pSim}{HTML}{6A45D8}
\definecolor{pRen}{HTML}{D9622B}
\definecolor{pAsk}{HTML}{C9A227}
\definecolor{pVal}{HTML}{2E9C8A}
\definecolor{pRel}{HTML}{5E7080}

\newcommand{\stage}[7]{%
  \fill[#5!7, draw=#5!70, line width=0.6pt, rounded corners=3pt] (#1,#2) rectangle (#3,#4);
  \begin{scope}
    \clip[rounded corners=3pt] (#1,#2) rectangle (#3,#4);
    \fill[#5!22] (#1,#4-0.52) rectangle (#3,#4);
  \end{scope}
  \draw[#5!70, line width=0.6pt, rounded corners=3pt] (#1,#2) rectangle (#3,#4);
  \fill[#5] (#1+0.3,#4-0.26) circle (0.17);
  \node[text=white, font=\sffamily\bfseries\scriptsize] at (#1+0.3,#4-0.26) {#6};
  \node[anchor=west, font=\sffamily\bfseries\footnotesize, text=black!85]
    at (#1+0.55,#4-0.26) {#7};
}

\stage{0}{0}{2.75}{3.9}{pSeed}{1}{Sample}
\node[body, text width=2.4cm] at (0.18,3.22)
  {One seeded parameter record per clip: scenario, object counts, colors, patterns.};
\node[anchor=north west, inner sep=3pt, fill=white, draw=pSeed!40, rounded corners=2pt,
      font=\ttfamily\tiny, align=left, text=black!75] at (0.18,1.62)
  {\{"scenario": "brake",\\ \ "seed": 1042,\\ \ "n\_gears": 6,\\ \ "driver": "teal"\}};

\stage{3.3}{0}{6.45}{3.9}{pSim}{2}{Simulate}
\node[body, text width=2.8cm] at (3.48,3.22)
  {Exact algebra or an event-driven simulation gives the scene state $s(t)$ at every $t$.};
\begin{scope}[shift={(3.6,0.35)}]
  \draw[black!35, line width=0.4pt] (0,0) -- (2.55,0);
  \draw[black!35, line width=0.4pt] (0,0) -- (0,1.25);
  \node[font=\sffamily\tiny, text=black!55, anchor=north east] at (2.55,0) {$t$};
  \draw[pSim, line width=0.9pt]
    (0.05,1.15) parabola (0.55,0)
    (0.55,0) parabola bend (0.98,0.8) (1.41,0)
    (1.41,0) parabola bend (1.74,0.46) (2.07,0)
    (2.07,0) parabola bend (2.3,0.22) (2.53,0);
  \foreach \x in {0.55,1.41,2.07}
    \fill[pSim] (\x,0) circle (0.05);
  \node[font=\sffamily\tiny, text=black!55, anchor=north east, align=right, inner sep=0pt]
    at (2.7,1.3) {scene checks:\\no overlaps,\\events apart};
\end{scope}

\stage{7.0}{2.05}{11.05}{3.9}{pRen}{3a}{Render}
\node[body, text width=2.35cm] at (7.18,3.22)
  {PIL draws each frame; ffmpeg encodes it. Only text: a clock.};
\begin{scope}[shift={(9.7,2.5)}, scale=0.85]
  \fill[black!80, rounded corners=1pt] (0,0) rectangle (1.25,0.95);
  \foreach \i in {0,...,4} {
    \fill[white] (0.08+\i*0.24,0.04) rectangle (0.2+\i*0.24,0.12);
    \fill[white] (0.08+\i*0.24,0.83) rectangle (0.2+\i*0.24,0.91);
  }
  \foreach \i/\c in {0/pRen!35, 1/pRen!55, 2/pRen!75} {
    \fill[\c] (0.07+\i*0.38,0.19) rectangle (0.41+\i*0.38,0.76);
    \fill[white] (0.17+\i*0.38+0.06*\i,0.47) circle (0.07);
  }
\end{scope}
\node[font=\sffamily\tiny, text=black!60] at (10.23,2.28) {720p, 24\,fps, $\le$\,10\,s};

\stage{7.0}{0}{11.05}{1.85}{pAsk}{3b}{Ask}
\node[body, text width=2.35cm] at (7.18,1.17)
  {Ten questions per clip, each with a derivation key and its params.};
\begin{scope}[shift={(9.6,0.2)}]
  \foreach \i/\h/\l in {0/0.43/L1, 1/0.60/L2, 2/0.60/L3, 3/0.43/L4, 4/0.43/L5} {
    \fill[pAsk!45!white] (\i*0.27,0.22) rectangle (\i*0.27+0.2,0.22+\h);
    \fill[pAsk] (\i*0.27,0.22) rectangle (\i*0.27+0.2,0.22+0.35*\h);
    \node[font=\sffamily\tiny, text=black!60] at (\i*0.27+0.1,0.1) {\l};
  }
\end{scope}

\stage{11.6}{0}{14.65}{3.9}{pVal}{4}{Validate}
\node[body, text width=2.75cm] at (11.78,3.22)
  {An independent answer module re-derives every answer from the stored scene.\\[3pt]
   Decoded frames are checked against $s(t)$; ladder, options, resolution, and duration are checked too.};
\node[anchor=south, fill=pVal, text=white, rounded corners=6pt, inner xsep=6pt, inner ysep=3pt,
      font=\sffamily\bfseries\scriptsize] at (13.125,0.22) {\ding{51}\ 0 mismatches};

\stage{15.2}{0}{17.25}{3.9}{pRel}{5}{Release}
\foreach \y/\t in {2.55/clip (.mp4), 1.7/final frame, 0.85/annotations} {
  \fill[white, draw=pRel!55, rounded corners=1.5pt] (15.42,\y) rectangle (17.03,\y+0.6);
  \fill[pRel!55] (15.42,\y) rectangle (15.55,\y+0.6);
  \node[anchor=west, font=\sffamily\scriptsize, text=black!80] at (15.62,\y+0.3) {\t};
}
\node[font=\sffamily\tiny, text=black!60, align=center] at (16.225,0.45)
  {final frame = control};

\draw[flow] (2.75,1.95) -- (3.3,1.95);
\draw[flow] (6.45,1.95) -- ++(0.25,0) |- (7.0,2.97);
\draw[flow] (6.45,1.95) -- ++(0.25,0) |- (7.0,0.92);
\draw[flow] (11.05,2.97) -- ++(0.25,0) |- (11.6,1.95);
\draw[flow] (11.05,0.92) -- ++(0.25,0) |- (11.6,1.95);
\draw[flow] (14.65,1.95) -- (15.2,1.95);
\draw[flow, pVal!80, dashed] (1.375,0) -- ++(0,-0.38) -| (12.4,0);
\node[fill=white, inner sep=1.5pt, font=\sffamily\tiny, text=pVal!90!black] at (8.6,-0.38)
  {stored params: the validator rebuilds the scene on its own};
\draw[flow, red!70!black, dashed] (13.85,0) -- ++(0,-0.72) -| (4.875,0);
\node[fill=white, inner sep=1.5pt, font=\sffamily\tiny, text=red!70!black] at (9.6,-0.72)
  {any mismatch: fix the generator, never loosen the check};
\end{tikzpicture}}

\caption{\textbf{Generation and validation pipeline.}
(1)~Each clip starts from a seeded parameter record.
(2)~Analytical expressions or deterministic simulations produce the
scene state $s(t)$, which drives both (3a)~the renderer and
(3b)~the ten-question ladder (bars: expected questions per level).
(4)~The validator reconstructs the scene from stored
parameters, re-derives the answers, and checks sampled decoded frames
against the expected state.
(5)~Each released clip includes its annotations and a held final frame
for static-image control.}
\label{fig:pipeline}
\end{figure*}

\paragraph{Consistency validation.}
For each sub-benchmark, a validator rebuilds every clip from its seed and
re-derives every answer from the drawn scene (\cref{fig:pipeline}). Where
possible it uses a different method from the generator, such as a
topological flood sweep, a sparse replay of cellular-automaton rules, or
brute-force face counting for spinning solids. In 26 sub-benchmarks it also
locates objects, flashes, and overlaps in decoded frames and compares them
with the expected state. During development these checks caught palette
collisions, drawing-order errors, and rounding errors, which we fixed in the
generators. On the release, the validators find zero mismatches in all 40
sub-benchmarks and flag all 120 answers we corrupted on purpose
(\cref{sec:validation}).

\paragraph{Questions designed to require video.}
Questions ask about properties of a trajectory: event counts, order,
direction, what happened while an object was hidden, or the result of an
intervention. When a question could be answered from the final frame, the
generator asks about a temporal property instead, such as which shape
appeared \emph{first} or which tile moved the \emph{most}, and the final
frame hides running totals. Of all questions, 81.1\% carry the
motion flag: 33\% on L1 and 87--95\% on L2--L5. The share
is lowest in \emph{robot long jump} (44.9\%) and highest in \emph{moving
shadows} (97.4\%).

\subsection{Reasoning families}
\label{sec:families}

\bench groups its 40 sub-benchmarks into seven reasoning families
(\cref{tab:families}), covering challenges relevant to robotics and
physical AI: predicting contact and motion, handling fluids, recovering
3D structure from viewpoint changes, tracking occluded objects,
monitoring evolving processes, reasoning about other agents, and
verifying rule-governed procedures. The complete list of
sub-benchmarks and their reasoning requirements is provided in
the supplementary material (\cref{tab:catalogue}).

\begin{table}[t]
\caption{\textbf{\bench composition and relevance to robotics and physical
AI.} \emph{Sub.}: sub-benchmarks ($2{,}500$ clips and $25{,}000$ questions
each). \emph{Var.}: scenario variants. \emph{V}: sub-benchmarks with injected
violations.}
\label{tab:families}
\centering
\scriptsize
\setlength{\tabcolsep}{2.5pt}
\begin{tabular}{@{}lrrr>{\raggedright\arraybackslash}p{0.47\linewidth}@{}}
\toprule
\textbf{Family} & \textbf{Sub.} & \textbf{Var.} & \textbf{V} & \textbf{Robotics counterpart}\\
\midrule
\rowcolor{rowgrey}
Mechanics \& motion        & 9 & 42 & 0 & Predicting drops, pushes, collisions, and stacking before acting\\
Fluids \& optics           & 6 & 28 & 1 & Pouring, filling, and draining; following light paths\\
\rowcolor{rowgrey}
Spatial \& 3D geometry     & 4 & 19 & 1 & Shape and pose from a moving camera or in-hand rotation\\
Occlusion \& permanence    & 6 & 25 & 0 & Tracking objects in clutter and behind the robot's arm\\
\rowcolor{rowgrey}
Temporal events            & 5 & 24 & 0 & Monitoring cascades; inferring a hidden rule\\
Agents \& navigation       & 5 & 23 & 0 & Inferring agent roles; traffic queues for driving\\
\rowcolor{rowgrey}
Rules \& algorithms        & 5 & 24 & 1 & Verifying a multi-step procedure and finding the faulty step\\
\midrule
\textbf{Total}             & \textbf{40} & \textbf{185} & \textbf{3} & $100{,}000$ clips, $1{,}000{,}000$ questions\\
\bottomrule
\end{tabular}
\end{table}

\subsection{Scenario variants and scene complexity}
\label{sec:variants}

A sub-benchmark is not a single scene template. Each generator samples one
of four or five \emph{scenario variants} per clip, totaling 185 across the
40 sub-benchmarks. These variants introduce changes in dynamics, event
sequences, geometry, or rules, beyond visual appearance alone. Each clip
also re-samples task-specific continuous and discrete parameters, including
object counts, sizes, speeds, event times, topology, colors, textures,
backgrounds, and camera settings. Of the one million questions,
$348{,}138$ are distinct question--answer pairs within their sub-benchmark,
from $1{,}007$ in \emph{rainbow bagel}, whose questions reuse a small set of
geometric answers, to $16{,}665$ in \emph{cellular automata}; repeated pairs
still come from different clips, so the answer must be read from each video
(details in \cref{tab:complexity,tab:spotlight}).

\subsection{The temporal reasoning ladder}
\label{sec:ladder}

Following the image-based GRIP benchmark~\cite{jawaid2026grip}, every clip
includes a five-rung ladder of questions designed to probe increasing
compositional depth: \textbf{L1} direct observation (\eg, ``How many gears
are in this train?''), \textbf{L2} relational reasoning (``Is the teal gear
rotating clockwise?''), \textbf{L3} measurement (``Which gear is the
slowest?''), \textbf{L4} multi-step integration (``In which round did a
piece move illegally?''), and \textbf{L5} counterfactual reasoning
(``If the driver gear reversed, which gears would also reverse?'').
Each clip has exactly ten questions: 1--3 on L1, L4, and L5 and 2--3 on
L2 and L3, about 17\% and 24\% of all questions per rung
(\cref{tab:levels}). Answer options are
shuffled per question, yielding nearly balanced correct letters
(A--D: 25.1, 25.1, 25.0, 24.9\%).
Each question stores a derivation key and its parameters, from which the
validator re-derives the answer. Counterfactual answers
come from fully re-simulating the altered scene, never from a local patch.
Every question is multiple choice with up to four options and a typed
ground truth.

\subsection{Injected violations}
\label{sec:violations}

In 3 sub-benchmarks, a controlled fraction of clips contains exactly one
rule violation that is \emph{not visually flagged} when it happens. In the
chess dances (about 20\% of clips), one piece makes an illegal move, either
the wrong pattern for its type or sliding through another piece. In the
flooded depth grid (about 20\%), one section either starts filling before any
full higher neighbor could feed it or receives water from a lower neighbor.
In the symmetry pictures (about 45\%), exactly one shape differs from its
orbit in fill, color, outline, size, turn, position, or trace direction. The
generator keeps each violation locally plausible, and subsequent states
follow the task rules from the altered state, keeping the rest of the clip
internally consistent. Detecting these violations requires checking the
observed events or configurations against the underlying rules. Questions
then ask whether a violation occurred, which object or section it involved,
and what a counterfactual correction would change downstream; ``none'' is
a valid answer in violation-free clips.

\section{Experimental Setup}
\label{sec:setup}

\paragraph{Models.}
We evaluate eleven open-weight VLMs (\cref{tab:overall}). Four run on the
Aleph/Vulcan inference service: Qwen3.6-27B, Qwen3.8-27B, the
Gemma-4-26B-A4B mixture-of-experts model, and Molmo2-8B. Three run on the
NRP Nautilus service: Qwen3.8-Flash-Next and two 4-bit quantization-aware
Gemma~4 builds (31B and 12B). We serve the remaining four with vLLM on
one or two H100 GPUs: Nemotron-3.5-Super-VL-120B-A12B, Keye-VL-1.5-8B,
the video-specialized InternVideo2.5-Chat-8B, and Reka Edge 2603.
Serving details and the evaluation subset are described in the supplementary material (\cref{sec:access}).

\begin{table*}[t]

\caption{\textbf{Evaluated models and overall accuracy} (\%, video condition,
100{,}000 questions attempted per model). \emph{Release}: month the model's
official Hugging Face repository was created. \emph{Params}: published parameter
count (total/active for mixture-of-experts models).
\emph{Weights}: served precision ($^\dagger$: precision of the official checkpoint; the service does not report it).
\emph{Strict}: accuracy requiring a leading option letter;
\emph{95\% CI}: clip-bootstrap interval;
\emph{Lenient}: accuracy with relaxed response parsing;
\emph{Adjusted}: strict accuracy adjusted against the constant-answer
baseline within each sub-benchmark and reasoning level;
\emph{Unparsed}: responses without a valid leading letter, scored as wrong.
Bold: best.}
\label{tab:overall}
\centering
\small
\setlength{\tabcolsep}{6pt}
\begin{tabular}{@{}llllrcrrr@{}}
\toprule
\textbf{Model} & \textbf{Release} & \textbf{Params} & \textbf{Weights} & \textbf{Strict} & \textbf{95\% CI} & \textbf{Lenient} & \textbf{Adjusted} & \textbf{Unparsed}\\
\midrule
\rowcolor{rowgrey}
Qwen3.6-27B & Apr 2026 & 27B & BF16 & \textbf{55.6} & 55.3--55.9 & \textbf{55.6} & \textbf{38.2} & 1.2\\
Qwen3.8-27B & Aug 2026 & 27B & FP8 & 49.7 & 49.4--50.0 & 49.7 & 30.0 & 5.8\\
\rowcolor{rowgrey}
Qwen3.8-Flash-Next & Aug 2026 & 180B/6B & FP8 & 49.6 & 49.3--49.9 & 49.6 & 29.9 & 5.1\\
Molmo2-8B & Dec 2025 & 8B & FP32$^\dagger$ & 46.8 & 46.5--47.1 & 46.8 & 26.0 & 0.0\\
\rowcolor{rowgrey}
Nemotron-3.5-Super-VL & Oct 2026 & 120B/12B & NVFP4 & 45.6 & 45.3--45.9 & 45.6 & 24.2 & 0.0\\
Gemma-4-31B & Mar 2026 & 31B & W4A16 & 44.8 & 44.6--45.2 & 45.1 & 23.2 & 15.7\\
\rowcolor{rowgrey}
Gemma-4-26B-A4B & Mar 2026 & 25.2B/3.8B & FP8 & 44.3 & 44.0--44.6 & 44.3 & 22.5 & 2.3\\
Keye-VL-1.5-8B & Aug 2025 & 8B & BF16 & 41.3 & 41.0--41.6 & 41.3 & 18.2 & 0.0\\
\rowcolor{rowgrey}
Gemma-4-12B & May 2026 & 12B & W4A16 & 38.8 & 38.5--39.1 & 39.7 & 14.8 & 7.1\\
InternVideo2.5-8B & Jan 2025 & 8B & BF16 & 36.9 & 36.6--37.2 & 36.9 & 12.2 & 0.0\\
\rowcolor{rowgrey}
Reka-Edge-2603 & Mar 2026 & 7B & BF16 & 31.2 & 30.9--31.5 & 31.2 & 4.3 & 0.0\\
\midrule
Chance (1/\#options) & & & & 25.6 & -- & 25.6 & -- & --\\
Constant answer & & & & 28.1 & -- & 28.1 & 0.0 & --\\
 
\bottomrule
\end{tabular}
\end{table*}

\paragraph{Protocol.}
Each question is a separate request containing the full MP4 clip as one
base64-encoded video, followed by the question, its
lettered options, and the instruction ``Answer with the letter of the correct
option only.'' On the two hosted services, frames are sampled by the server
rather than our client. On our own servers, Nemotron and Keye sample at
2\,fps (with a configured maximum of 256 frames), InternVideo2.5 uses
32 frames, and Reka Edge uses 16 frames, so frame sampling differs between models. Decoding is greedy (temperature~0)
with at most 16 output tokens, and thinking is disabled where supported.
No additional system prompt is used, except that InternVideo2.5 retains
its checkpoint's system message.
The exact prompt is in \cref{sec:prompts}.

\paragraph{Comparability.}
Serving stack, precision, video sampling, thinking control, and system
prompt differ between models (\cref{tab:conditions}), so each result measures
a model under its serving configuration. The ranking is identical under lenient scoring (Kendall's
$\tau=1.0$) and largely preserved on parsed replies ($\tau=0.82$;
\cref{tab:rank_robust}).

\paragraph{Input conditions.}
The main evaluation gives each model the full clip. Two controls use 10{,}000
questions (25 clips per sub-benchmark): \textbf{final-frame} shows only the
last frame (five models), and \textbf{shuffled} shows the same 16 frames in
order or in random order (three models). The accuracy differences are the
\emph{motion gain} and the \emph{order gain} (\cref{sec:metric_details}).

\paragraph{Metrics.}
A reply is scored \emph{strictly}: it must start with an option letter, and
an unparsed reply counts as wrong. We also report a \emph{lenient} score that
accepts ``the answer is X'' or a letter after leading punctuation. To account
for answer-letter imbalance, we report \emph{baseline-adjusted accuracy}
$a_{\mathrm{adj}} = (a_{\mathrm{raw}} - b)/(1-b)$, as in
GRIP~\cite{jawaid2026grip}. Here $b$ is the constant-answer baseline,
defined by the most frequent correct letter within each sub-benchmark and
ladder level. Adjustment is applied within these groups before aggregation.
Zero indicates baseline performance, and one indicates perfect accuracy.
The overall baseline is 28.1\%, compared with 25.6\% expected from random
guessing. Each sub-benchmark contributes 2,500 evaluation questions, so
pooled and sub-benchmark-averaged accuracies are nearly identical. Family scores average the
sub-benchmarks within each family. We compute 95\% confidence intervals
from 1,000 bootstrap resamples of clips within each sub-benchmark.

\section{Results}
\label{sec:results}


\paragraph{\bench is far from solved.}
\cref{tab:overall} shows that the best-performing model,
Qwen3.6-27B, answers 55.6\% of the 100{,}000 questions, while the weakest,
Reka-Edge-2603, achieves 31.2\%, only 3.1 percentage points above the
constant-answer baseline of 28.1\%. Baseline-adjusted accuracy ranges
from 4.3 to 38.2 points, so even the strongest model closes only about
two fifths of the gap between the baseline and perfect accuracy; the 95\%
intervals are about $\pm0.3$ points.

\paragraph{Accuracy generally declines along the ladder.}
\cref{fig:ladder} and \cref{tab:ladder} in the supplementary material break down accuracy by reasoning level.
Every model performs best on L1 (direct observation) or L2, and nine
of eleven perform worst on L5 (counterfactual reasoning). Raw accuracy
drops by up to 37 points from L1 to L5. After baseline adjustment,
Nemotron-3.5-Super-VL retains 16.0 points on L5, Qwen3.6-27B 15.4,
and Molmo2-8B 13.2, while the remaining eight models score between
$-1.2$ and 6.0 points.

\begin{figure}[t]
\centering
\includegraphics[width=\linewidth]{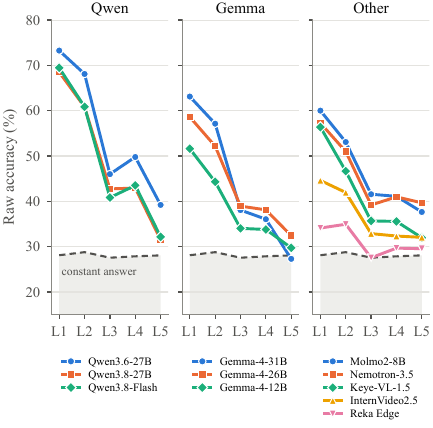}

\caption{\textbf{Raw accuracy across the five-rung ladder} (\%, strict),
one panel per model family. Dashed line and grey band: the constant-answer
baseline of each rung.
The largest raw L5 advantage over the baseline is 11.6 points.}
\label{fig:ladder}
\end{figure}

\begin{figure*}[t]
\centering
\includegraphics[width=\linewidth]{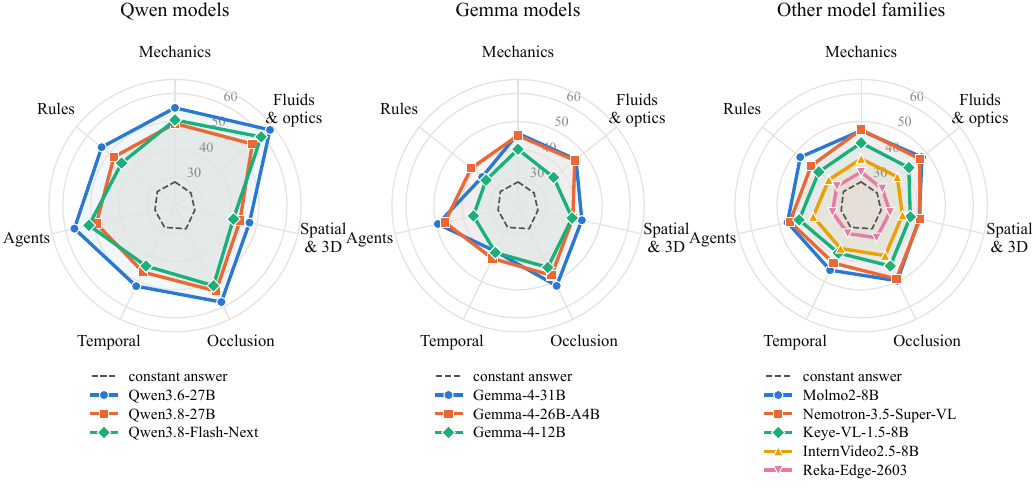}
\caption{\textbf{Accuracy by reasoning family} (\%, strict; each family is
the mean of its sub-benchmarks). Dashed heptagon: the constant-answer
baseline. Qwen, Gemma, and the five other model families are drawn in
separate panels so that no more than five profiles overlap. The values are listed in \cref{tab:families_full}.}
\label{fig:radar}
\end{figure*}

\paragraph{Families: occlusion and fluids are easiest, spatial geometry is hardest.}
\cref{fig:radar} shows that, averaged over the eleven
models, occlusion \& permanence (47.5\%) and fluids \& optics (46.3\%)
have the highest accuracy, while spatial \& 3D geometry is the hardest
(40.2\%; 30.6--47.3\% across models). Temporal events and rule tracking
also reveal differences between model families: the three Gemma models
score 34.6--41.4\%, compared with 43.9--53.5\% for the three Qwen models
and 45.5--47.8\% for Molmo2-8B.

\begin{table}[t]
\caption{\textbf{Final-frame control} (baseline-adjusted accuracy, points;
strict; 25 clips per sub-benchmark, 10{,}000 questions per model).
\emph{Gain}: video minus final frame, with a 95\% clip-bootstrap interval;
\emph{Flag.}/\emph{Unfl.}: gain with and without the motion flag.
InternVideo2.5 sees the image without its video token compression.}
\label{tab:final_frame}
\centering
\scriptsize
\setlength{\tabcolsep}{3.5pt}
\begin{tabular}{@{}lrrcrr@{}}
\toprule
\textbf{Model} & \textbf{Video} & \textbf{Final} & \textbf{Gain [95\% CI]} & \textbf{Flag.} & \textbf{Unfl.}\\
\midrule
\rowcolor{rowgrey}
Gemma-4-26B-A4B & 22.3 & 7.6 & 14.7 [13.2, 16.2] & 17.3 & 7.0\\
Qwen3.8-27B & 30.2 & 15.6 & 14.5 [13.1, 15.8] & 17.9 & 11.3\\
\rowcolor{rowgrey}
Gemma-4-31B & 22.9 & 8.8 & 14.1 [12.6, 15.5] & 19.7 & 3.0\\
Molmo2-8B & 26.0 & 19.8 & 6.2 [4.7, 7.6] & 10.5 & $-$0.6\\
\rowcolor{rowgrey}
InternVideo2.5-8B & 11.8 & 10.2 & 1.7 [0.3, 3.1] & 4.3 & $-$5.1\\
 
\bottomrule
\end{tabular}
\end{table}

\begin{figure*}[t]
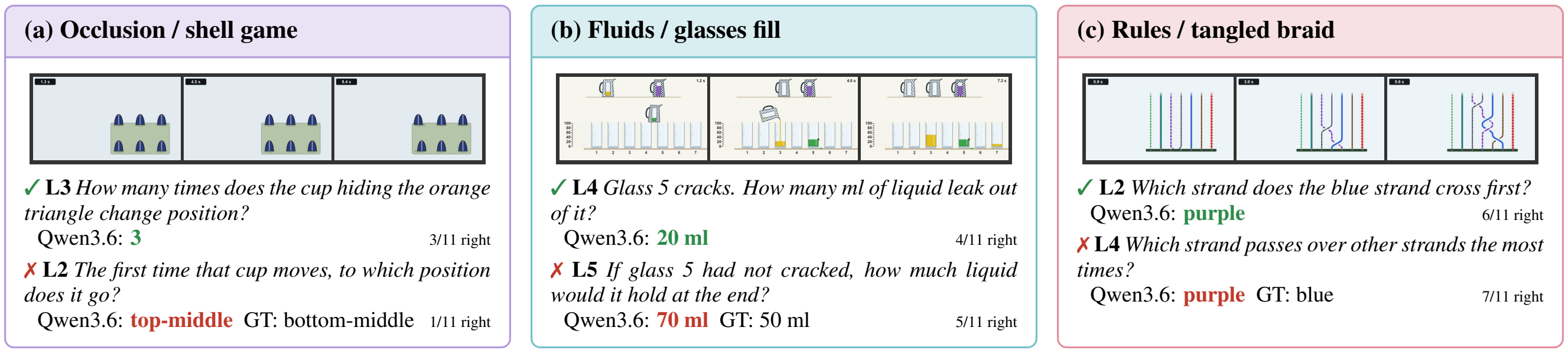

\centering
\begin{minipage}[t]{0.325\textwidth}
\begin{tcolorbox}[enhanced, equal height group=qual, colback=white, colframe=famOccl!65!white,
  colbacktitle=famOccl!18!white, coltitle=black, boxrule=0.6pt, arc=2.5pt, left=3pt, right=3pt,
  top=2pt, bottom=2pt, toptitle=1.5pt, bottomtitle=1.5pt, fonttitle=\footnotesize\bfseries,
  title={(a) Occlusion / shell game}]
\filmstrip{shell_game_tracking}
{\scriptsize
{\color{okgreen}\cmark}~\textbf{L3} \emph{How many times does the cup hiding the orange triangle change position?}\par
\hspace*{0.6em}Qwen3.6: \textbf{\color{okgreen}3} \hfill {\tiny 3/11 right}\par
\vspace{2pt}
{\color{badred}\xmark}~\textbf{L2} \emph{The first time that cup moves, to which position does it go?}\par
\hspace*{0.6em}Qwen3.6: \textbf{\color{badred}top-middle}\enspace GT: bottom-middle \hfill {\tiny 1/11 right}\par}
\end{tcolorbox}
\end{minipage}\hfill
\begin{minipage}[t]{0.325\textwidth}
\begin{tcolorbox}[enhanced, equal height group=qual, colback=white, colframe=famFluid!65!white,
  colbacktitle=famFluid!18!white, coltitle=black, boxrule=0.6pt, arc=2.5pt, left=3pt, right=3pt,
  top=2pt, bottom=2pt, toptitle=1.5pt, bottomtitle=1.5pt, fonttitle=\footnotesize\bfseries,
  title={(b) Fluids / glasses fill}]
\filmstrip{glasses_fill}
{\scriptsize
{\color{okgreen}\cmark}~\textbf{L4} \emph{Glass 5 cracks. How many ml of liquid leak out of it?}\par
\hspace*{0.6em}Qwen3.6: \textbf{\color{okgreen}20 ml} \hfill {\tiny 4/11 right}\par
\vspace{2pt}
{\color{badred}\xmark}~\textbf{L5} \emph{If glass 5 had not cracked, how much liquid would it hold at the end?}\par
\hspace*{0.6em}Qwen3.6: \textbf{\color{badred}70 ml}\enspace GT: 50 ml \hfill {\tiny 5/11 right}\par}
\end{tcolorbox}
\end{minipage}\hfill
\begin{minipage}[t]{0.325\textwidth}
\begin{tcolorbox}[enhanced, equal height group=qual, colback=white, colframe=famRules!65!white,
  colbacktitle=famRules!18!white, coltitle=black, boxrule=0.6pt, arc=2.5pt, left=3pt, right=3pt,
  top=2pt, bottom=2pt, toptitle=1.5pt, bottomtitle=1.5pt, fonttitle=\footnotesize\bfseries,
  title={(c) Rules / tangled braid}]
\filmstrip{tangled_braid}
{\scriptsize
{\color{okgreen}\cmark}~\textbf{L2} \emph{Which strand does the blue strand cross first?}\par
\hspace*{0.6em}Qwen3.6: \textbf{\color{okgreen}purple} \hfill {\tiny 6/11 right}\par
\vspace{2pt}
{\color{badred}\xmark}~\textbf{L4} \emph{Which strand passes over other strands the most times?}\par
\hspace*{0.6em}Qwen3.6: \textbf{\color{badred}purple}\enspace GT: blue \hfill {\tiny 7/11 right}\par}
\end{tcolorbox}
\end{minipage}
\caption{\textbf{Qualitative results: the best model on three clips}
(three frames each, left to right in time). For each clip, one question that Qwen3.6-27B answers correctly (\cmark) and one
that it gets wrong (\xmark), with how many of the eleven models answer it
correctly. The model counts how often the cup moves but places it wrongly after
its first move (a), measures the leak but cannot undo it in the counterfactual
(b), and finds the first crossing but not which strand passes over most (c).}
\label{fig:qualitative}
\end{figure*}

\paragraph{Models use motion, but flagged questions remain harder.}
Every model above 40\% scores 11.2--19.1 points lower on questions with the
motion flag (\cref{tab:motion}). Part of this gap comes from difficulty,
because flagged questions sit mostly on higher rungs; within the same
sub-benchmark and rung the gap is 3.8--8.0 points (\cref{tab:motion_strat}).
The controls show that the stronger models do use the motion
(\cref{tab:final_frame}). With only the last frame, Qwen3.8-27B and both Gemma
models lose 14.1--14.7 points of adjusted accuracy, and on flagged questions
the Gemma models fall to the baseline. The loss is larger on flagged questions
for every model, which supports the flag. Shuffling the frames costs
Qwen3.8-Flash-Next and Gemma-4-31B about 10 points and Gemma-4-12B 1.5
(\cref{tab:shuffled}). Some Gemma replies refuse to answer from a single image.
On single clips, the best model reads counts and first events but loses
track of where an object went and fails counterfactuals (\cref{fig:qualitative}).

\paragraph{Violation questions are not solved reliably.}
Detection F1 stays between 23.0 and 47.2, and localization accuracy between
5.2\% and 40.2\% (\cref{tab:violation} in the supplementary material).
Qwen3.6-27B, Molmo2-8B, and Keye-VL-1.5-8B almost always answer ``nothing
went wrong'': they are right on 83--87\% of clean clips but find none of the
25 illegal chess moves. Qwen3.8-Flash-Next and Gemma-4-26B-A4B rarely do,
which gives them the best F1 but only 14.2\% and 18.1\% on clean clips.

\paragraph{Implications for robotics and physical AI.}
A planner needs counterfactuals and an execution monitor needs violation
detection, and these are the weakest results, so current VLMs cannot yet
serve as the video monitors of recent robot
work~\cite{chan2026online,chen2026topreward}.

\paragraph{Human baseline.}
Three annotators answered $7{,}720$ of the evaluated questions from all seven
families and got 91.2\% right, against 61.1\% for the best model on the same
questions (supplementary \cref{sec:human}).

\section{Limitations}
\label{sec:limitations}

\bench prioritizes control over ecological realism. Its clips are clean,
mostly two-dimensional, and rendered on plain backgrounds, so performance on it
should not be read as a direct estimate of real-world video understanding.
It does not test closed-loop control, real camera noise, or embodiment.
The results cover eleven open-weight models on a 10\% subset, in one
non-thinking run each, with serving stacks, precision, and frame budgets
(16 frames to 256) that differ between models. The final-frame and shuffled
controls cover five and three models. Finally, a few
conventions, such as a pour delivering only what is left in a jug or how
flood inflow is split among lower neighbors, are modeling choices that we
document rather than physical necessities.

\newpage
\section{Conclusion}
\label{sec:conclusion}

We introduced \bench, 40 procedurally generated video sub-benchmarks
with one million questions and state-derived ground truth. The best of
eleven open-weight VLMs answers 55.6\% of the questions. Models do use
motion, losing 14--15 adjusted points with only the final frame, but they
stay near the baseline on counterfactual questions and trade missed
violations against false alarms. With 185 scenario variants and validated
answers, \bench is a controlled test of the temporal, geometric, and
physical reasoning that embodied AI needs, and a source of procedurally
generated training data.

\section*{LLM Usage Statement}
During the preparation of this work, the authors used Claude Sonnet and ChatGPT to improve grammar and wording. All final text was reviewed and approved by the authors, who maintain full responsibility for the content of the manuscript.

\clearpage
{
    \small
    \bibliographystyle{ieeenat_fullname}
    \bibliography{main}

\begin{thebibliography}{38}
\providecommand{\natexlab}[1]{#1}
\providecommand{\url}[1]{\texttt{#1}}
\expandafter\ifx\csname urlstyle\endcsname\relax
  \providecommand{\doi}[1]{doi: #1}\else
  \providecommand{\doi}{doi: \begingroup \urlstyle{rm}\Url}\fi

\bibitem[Barrett et~al.(2018)Barrett, Hill, Santoro, Morcos, and
  Lillicrap]{barrett2018measuring}
David G.~T. Barrett, Felix Hill, Adam Santoro, Ari~S. Morcos, and Timothy
  Lillicrap.
\newblock Measuring abstract reasoning in neural networks.
\newblock In \emph{International Conference on Machine Learning}, 2018.

\bibitem[Bear et~al.(2021)]{bear2021physion}
Daniel~M. Bear et~al.
\newblock Physion: Evaluating physical prediction from vision in humans and
  machines.
\newblock In \emph{NeurIPS Datasets and Benchmarks Track}, 2021.

\bibitem[Brohan et~al.(2023)]{brohan2023rt2}
Anthony Brohan et~al.
\newblock {RT-2}: Vision-language-action models transfer web knowledge to
  robotic control.
\newblock In \emph{Conference on Robot Learning}, 2023.

\bibitem[Chan et~al.(2026)]{chan2026online}
Yung-Shun Chan et~al.
\newblock Online robot manipulation error detection and description.
\newblock In \emph{Conference on Robot Learning}, 2026.

\bibitem[Chen et~al.(2021)]{chen2021geoqa}
Jiaqi Chen et~al.
\newblock {GeoQA}: A geometric question answering benchmark towards multimodal
  numerical reasoning.
\newblock In \emph{Findings of ACL}, 2021.

\bibitem[Chen et~al.(2026)]{chen2026topreward}
Shirui Chen et~al.
\newblock {TOPReward}: Token probabilities as hidden zero-shot rewards for
  robotics.
\newblock In \emph{Conference on Robot Learning}, 2026.

\bibitem[Chow et~al.(2025)]{chow2025physbench}
Wei Chow et~al.
\newblock {PhysBench}: Benchmarking and enhancing vision-language models for
  physical world understanding.
\newblock In \emph{International Conference on Learning Representations}, 2025.
\newblock arXiv:2501.16411.

\bibitem[Driess et~al.(2023)]{driess2023palme}
Danny Driess et~al.
\newblock {PaLM-E}: An embodied multimodal language model.
\newblock In \emph{International Conference on Machine Learning}, 2023.

\bibitem[Fang et~al.(2026)]{fang2026molmoact2}
Haoquan Fang et~al.
\newblock {MolmoAct2}: Action reasoning models for real-world deployment.
\newblock In \emph{Conference on Robot Learning}, 2026.

\bibitem[Feng et~al.(2026)]{feng2026tempo}
Zhenyang Feng et~al.
\newblock {TEMPO}: Learning temporal context for dynamic robot manipulation.
\newblock In \emph{Conference on Robot Learning}, 2026.

\bibitem[Fu et~al.(2025)Fu, Dai, Luo, Li, Ren, Zhang, Wang, Zhou, Shen, Zhang,
  et~al.]{fu2025videomme}
Chaoyou Fu, Yuhan Dai, Yongdong Luo, Lei Li, Shuhuai Ren, Renrui Zhang, Zihan
  Wang, Chenyu Zhou, Yunhang Shen, Mengdan Zhang, et~al.
\newblock {Video-MME}: The first-ever comprehensive evaluation benchmark of
  multi-modal {LLMs} in video analysis.
\newblock In \emph{CVPR}, 2025.

\bibitem[Girdhar and Ramanan(2020)]{girdhar2020cater}
Rohit Girdhar and Deva Ramanan.
\newblock {CATER}: A diagnostic dataset for compositional actions and
  {TE}mporal reasoning.
\newblock In \emph{ICLR}, 2020.

\bibitem[Jawaid et~al.(2026)Jawaid, Hanif, Uddin, Zafar, Alqurashi, Zafar,
  Farooqui, Khan, Sadak, Raza, and Qureshi]{jawaid2026grip}
Bilal Jawaid, Hamza Hanif, Muhammad~Maaz Uddin, Anas Zafar, Inad Alqurashi,
  Aizan Zafar, Osama Farooqui, Sheheryar Khan, Ferhat Sadak, Shaina Raza, and
  Rizwan Qureshi.
\newblock Geometric parrots: Rule absence vs. rule fragility in spatial and
  physical reasoning for vision-language models.
\newblock \emph{arXiv preprint arXiv:2610.16170}, 2026.

\bibitem[Kim et~al.(2024)]{kim2024openvla}
Moo~Jin Kim et~al.
\newblock {OpenVLA}: An open-source vision-language-action model.
\newblock In \emph{Conference on Robot Learning}, 2024.

\bibitem[Lerer et~al.(2016)Lerer, Gross, and Fergus]{lerer2016blocktowers}
Adam Lerer, Sam Gross, and Rob Fergus.
\newblock Learning physical intuition of block towers by example.
\newblock In \emph{International Conference on Machine Learning}, 2016.

\bibitem[Li et~al.(2024)Li, Wang, He, Li, Wang, Liu, Wang, Xu, Chen, Luo, Wang,
  and Qiao]{li2024mvbench}
Kunchang Li, Yali Wang, Yinan He, Yizhuo Li, Yi Wang, Yi Liu, Zun Wang, Jilan
  Xu, Guo Chen, Ping Luo, Limin Wang, and Yu Qiao.
\newblock {MVBench}: A comprehensive multi-modal video understanding benchmark.
\newblock In \emph{CVPR}, 2024.

\bibitem[Liu and Lee(2026)]{liu2026vetbench}
Tiedong Liu and Wee~Sun Lee.
\newblock Can vision-language models solve the shell game?
\newblock \emph{arXiv preprint arXiv:2603.08436}, 2026.

\bibitem[Liu et~al.(2024)Liu, Li, Liu, Wang, Ren, Li, Chen, Sun, and
  Hou]{liu2024tempcompass}
Yuanxin Liu, Shicheng Li, Yi Liu, Yuxiang Wang, Shuhuai Ren, Lei Li, Sishuo
  Chen, Xu Sun, and Lu Hou.
\newblock {TempCompass}: Do video {LLMs} really understand videos?
\newblock In \emph{Findings of ACL}, 2024.

\bibitem[Liu et~al.(2026)Liu, Ouyang, Wu, Liu, Sui, Li, Zhong, Charles, Zhou,
  and Sun]{liu2026videoreasonbench}
Yuanxin Liu, Kun Ouyang, Haoning Wu, Yi Liu, Lin Sui, Xinhao Li, Yan Zhong, Y.
  Charles, Xinyu Zhou, and Xu Sun.
\newblock {VideoReasonBench}: Can {MLLMs} perform vision-centric complex video
  reasoning?
\newblock In \emph{ICLR}, 2026.

\bibitem[Lu et~al.(2024)]{lu2024mathvista}
Pan Lu et~al.
\newblock {MathVista}: Evaluating mathematical reasoning of foundation models
  in visual contexts.
\newblock In \emph{International Conference on Learning Representations}, 2024.

\bibitem[Mangalam et~al.(2023)Mangalam, Akshulakov, and
  Malik]{mangalam2023egoschema}
Karttikeya Mangalam, Raiymbek Akshulakov, and Jitendra Malik.
\newblock {EgoSchema}: A diagnostic benchmark for very long-form video language
  understanding.
\newblock In \emph{NeurIPS}, 2023.

\bibitem[Maurya et~al.(2026)]{maurya2026imbench}
Anurag Maurya et~al.
\newblock {IMBench}: A benchmark for intuitive robotic manipulation.
\newblock In \emph{Conference on Robot Learning}, 2026.

\bibitem[{NVIDIA}(2025)]{nvidia2025cosmos}
{NVIDIA}.
\newblock Cosmos world foundation model platform for physical {AI}.
\newblock \emph{arXiv preprint arXiv:2501.03575}, 2025.

\bibitem[P{\u{a}}tr{\u{a}}ucean et~al.(2023)P{\u{a}}tr{\u{a}}ucean, Smaira,
  Gupta, Recasens, Markeeva, Banarse, Koppula, Heyward, Malinowski, Yang,
  et~al.]{patraucean2023perception}
Viorica P{\u{a}}tr{\u{a}}ucean, Lucas Smaira, Ankush Gupta, Adri{\`a} Recasens,
  Larisa Markeeva, Dylan Banarse, Skanda Koppula, Joseph Heyward, Mateusz
  Malinowski, Yi Yang, et~al.
\newblock Perception test: A diagnostic benchmark for multimodal video models.
\newblock In \emph{NeurIPS}, 2023.

\bibitem[Rahman et~al.(2025)Rahman, Anwar, Usman, Ahmad, and
  Mian]{rahman2025pendulum}
A.~B. M.~Ashikur Rahman, Saeed Anwar, Muhammad Usman, Irfan Ahmad, and Ajmal
  Mian.
\newblock {PENDULUM}: A benchmark for assessing sycophancy in multimodal large
  language models.
\newblock \emph{arXiv preprint arXiv:2512.19350}, 2025.

\bibitem[Rahmanzadehgervi et~al.(2024)Rahmanzadehgervi, Bolton, Taesiri, and
  Nguyen]{rahmanzadehgervi2024blind}
Pooyan Rahmanzadehgervi, Logan Bolton, Mohammad~Reza Taesiri, and Anh~Totti
  Nguyen.
\newblock Vision language models are blind: Failing to translate detailed
  visual features into words.
\newblock \emph{arXiv preprint arXiv:2407.06581}, 2024.

\bibitem[Riochet et~al.(2021)Riochet, Castro, Bernard, Lerer, Fergus, Izard,
  and Dupoux]{riochet2021intphys}
Ronan Riochet, Mario~Ynocente Castro, Mathieu Bernard, Adam Lerer, Rob Fergus,
  V{\'e}ronique Izard, and Emmanuel Dupoux.
\newblock {IntPhys} 2019: A benchmark for visual intuitive physics
  understanding.
\newblock \emph{IEEE TPAMI}, 44\penalty0 (9):\penalty0 5016--5025, 2021.

\bibitem[Schulze~Buschoff et~al.(2026)Schulze~Buschoff, Voudouris, Demircan,
  and Schulz]{schulzebuschoff2026interaction}
Luca~M. Schulze~Buschoff, Konstantinos Voudouris, Can Demircan, and Eric
  Schulz.
\newblock Can vision language models learn intuitive physics from interaction?
\newblock \emph{arXiv preprint arXiv:2602.06033}, 2026.

\bibitem[Shangguan et~al.(2025)Shangguan, Li, Ding, Zheng, Zhao, Fitzgerald,
  and Cohan]{shangguan2025tomato}
Ziyao Shangguan, Chuhan Li, Yuxuan Ding, Yanan Zheng, Yilun Zhao, Tesca
  Fitzgerald, and Arman Cohan.
\newblock {TOMATO}: Assessing visual temporal reasoning capabilities in
  multimodal foundation models.
\newblock In \emph{ICLR}, 2025.

\bibitem[Sharma et~al.(2024)Sharma, Tong, Korbak, Duvenaud, Askell, Bowman,
  et~al.]{sharma2024sycophancy}
Mrinank Sharma, Meg Tong, Tomasz Korbak, David Duvenaud, Amanda Askell,
  Samuel~R. Bowman, et~al.
\newblock Towards understanding sycophancy in language models.
\newblock In \emph{International Conference on Learning Representations}, 2024.
\newblock arXiv:2310.13548.

\bibitem[Wang et~al.(2026{\natexlab{a}})Wang, Xu, and Huang]{wang2026collision}
Jun Wang, Xiaohao Xu, and Xiaonan Huang.
\newblock Probing collision grounding in vision-language models for safe
  human--robot collaboration.
\newblock In \emph{Conference on Robot Learning}, 2026{\natexlab{a}}.

\bibitem[Wang et~al.(2026{\natexlab{b}})]{wang2026liquidlab}
Yuzhou Wang et~al.
\newblock {LiquidLab}: Benchmarking robotic automation for liquid operation in
  laboratory.
\newblock In \emph{Conference on Robot Learning}, 2026{\natexlab{b}}.

\bibitem[Yang et~al.(2026)]{yang2026eventvla}
Ganlin Yang et~al.
\newblock {EventVLA}: Event-driven visual evidence memory for long-horizon
  vision-language-action policies.
\newblock In \emph{Conference on Robot Learning}, 2026.

\bibitem[Yeom et~al.(2026)]{yeom2026latents}
Jewon Yeom et~al.
\newblock What makes video world model latents action-relevant: Prediction over
  reconstruction.
\newblock In \emph{Conference on Robot Learning}, 2026.

\bibitem[Yi et~al.(2020)Yi, Gan, Li, Wu, Kohli, et~al.]{yi2020clevrer}
Kexin Yi, Chuang Gan, Yunzhu Li, Pushmeet Wu, Pushmeet Kohli, et~al.
\newblock {CLEVRER}: Collision events for video representation and reasoning.
\newblock In \emph{International Conference on Learning Representations}, 2020.

\bibitem[Yu et~al.(2026)Yu, Ma, Huang, Lee, Yang, Choi, Brown, Michel, Zheng,
  Shin, and Xie]{yu2026vstat}
Sihyun Yu, Nanye Ma, Pinzhi Huang, Hyunseok Lee, Shusheng Yang, June~Suk Choi,
  Ellis Brown, Oscar Michel, Boyang Zheng, Jinwoo Shin, and Saining Xie.
\newblock Benchmarking visual state tracking in multimodal video understanding.
\newblock \emph{arXiv preprint arXiv:2606.03920}, 2026.

\bibitem[Zhang et~al.(2019)Zhang, Gao, Jia, Zhu, and Zhu]{zhang2019raven}
Chi Zhang, Feng Gao, Baoxiong Jia, Yixin Zhu, and Song-Chun Zhu.
\newblock {RAVEN}: A dataset for relational and analogical visual reasoning.
\newblock In \emph{IEEE/CVF Conference on Computer Vision and Pattern
  Recognition}, 2019.

\bibitem[Zhang et~al.(2024)]{zhang2024mathverse}
Renrui Zhang et~al.
\newblock {MathVerse}: Does your multi-modal {LLM} truly see the diagrams in
  visual math problems?
\newblock In \emph{European Conference on Computer Vision}, 2024.

\end{thebibliography}
}

\clearpage
\appendix
\setcounter{table}{0}
\setcounter{figure}{0}
\renewcommand{\thesection}{\Alph{section}}
\renewcommand{\thetable}{S\arabic{table}}
\renewcommand{\thefigure}{S\arabic{figure}}
\maketitlesupplementary

\section{Example clips}
\label{sec:examples}

\begin{figure*}[p]
\centering
\setlength{\tabcolsep}{1.5pt}
\renewcommand{\arraystretch}{0.6}
\begin{tabular}{@{}c *{3}{>{\centering\arraybackslash}m{0.215\linewidth}} >{\raggedright\arraybackslash}m{0.28\linewidth}@{}}
& {\scriptsize $t=0.15\,T$} & {\scriptsize $t=0.50\,T$} & {\scriptsize $t=0.92\,T$} & {\scriptsize Questions from the clip's ladder}\\[2pt]
\rotatebox{90}{\parbox{0.13\linewidth}{\centering\scriptsize Glasses fill}} &
\includegraphics[width=\linewidth]{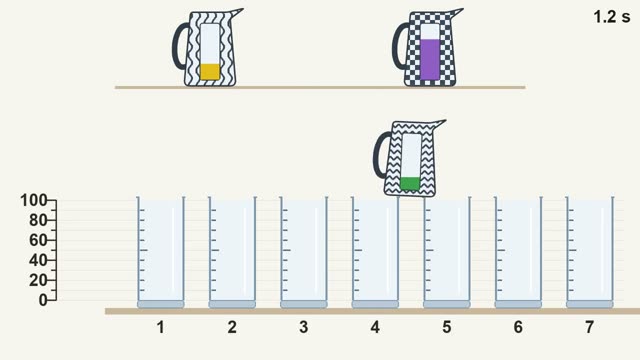} &
\includegraphics[width=\linewidth]{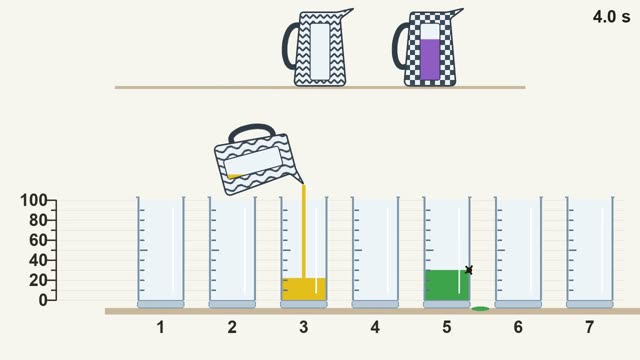} &
\includegraphics[width=\linewidth]{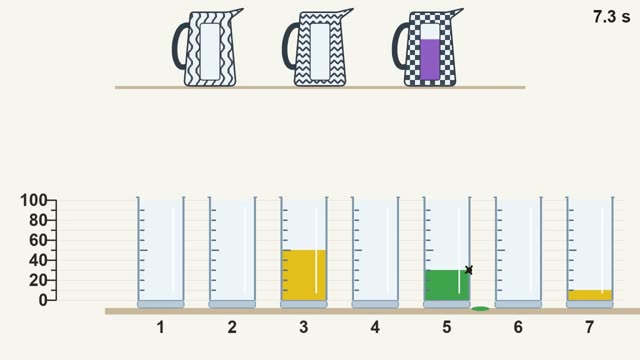} &
\parbox[c]{\linewidth}{\scriptsize\raggedright \textbf{L2:} \emph{Which glass is the last one to receive liquid in the video?} GT: glass 7.\\[1pt]
\textbf{L5:} \emph{If glass 5 had NOT cracked, how much liquid would it hold at the end of the video?} GT: 50 ml.}\\[2pt]
\rotatebox{90}{\parbox{0.13\linewidth}{\centering\scriptsize Buoyancy tank}} &
\includegraphics[width=\linewidth]{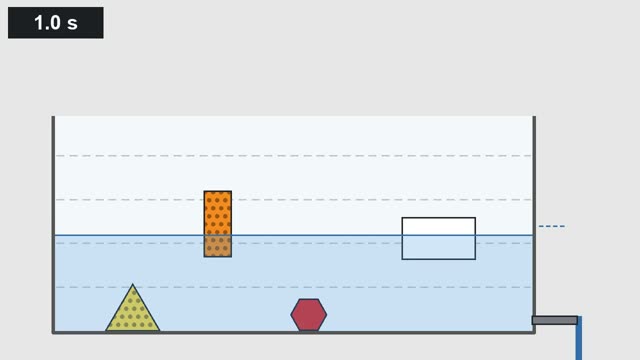} &
\includegraphics[width=\linewidth]{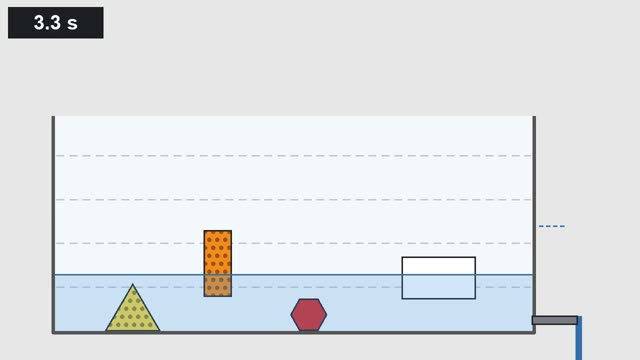} &
\includegraphics[width=\linewidth]{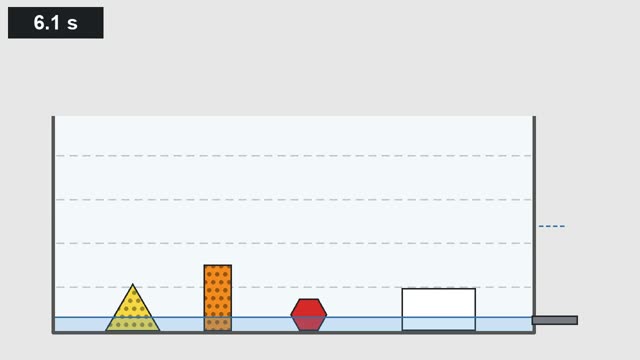} &
\parbox[c]{\linewidth}{\scriptsize\raggedright \textbf{L2:} \emph{At the end, whose top is higher above the tank floor: the red hexagon or the white wide block?} GT: the white wide block.\\[1pt]
\textbf{L5:} \emph{If the drain had been closed one second earlier, how many objects would end up resting on the tank floor?} GT: 2.}\\[2pt]
\rotatebox{90}{\parbox{0.13\linewidth}{\centering\scriptsize Well flood}} &
\includegraphics[width=\linewidth]{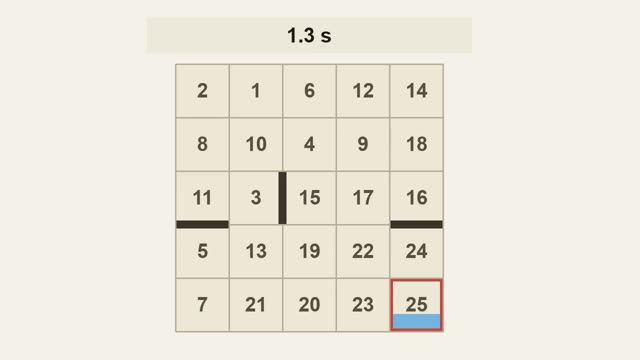} &
\includegraphics[width=\linewidth]{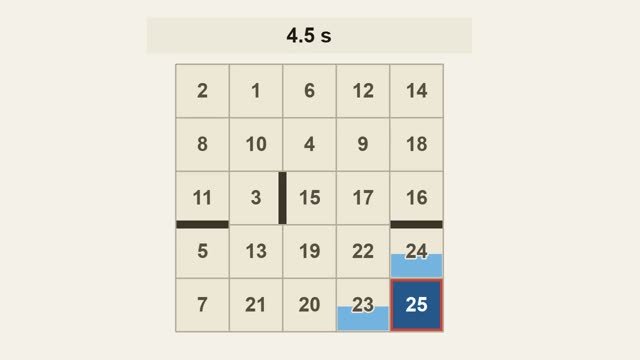} &
\includegraphics[width=\linewidth]{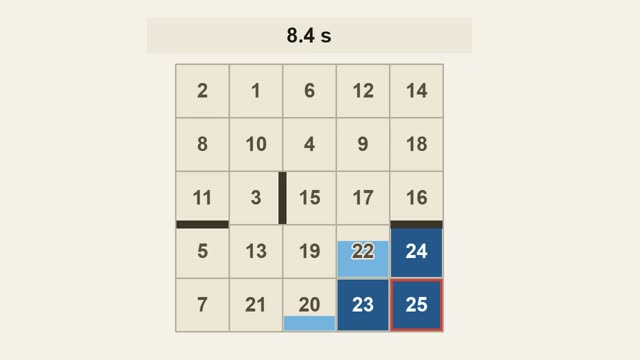} &
\parbox[c]{\linewidth}{\scriptsize\raggedright \textbf{L2:} \emph{Of these sections, which one did water reach FIRST?} GT: the section with depth value 25.\\[1pt]
\textbf{L5:} \emph{If the section with depth value 24 had been sealed so that no water could enter it, how many distinct sections would water have reached?} GT: 6.}\\[2pt]
\rotatebox{90}{\parbox{0.13\linewidth}{\centering\scriptsize Rolling races}} &
\includegraphics[width=\linewidth]{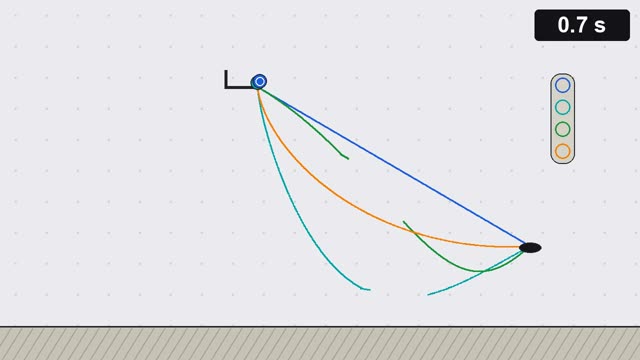} &
\includegraphics[width=\linewidth]{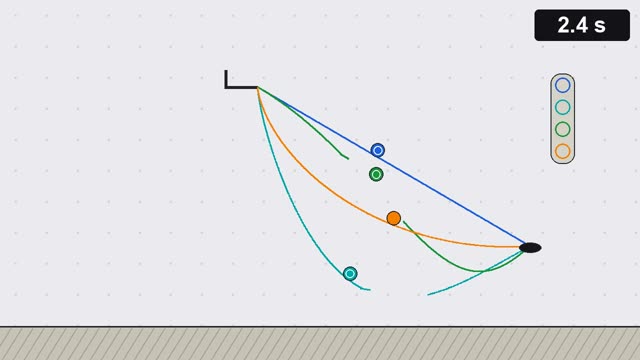} &
\includegraphics[width=\linewidth]{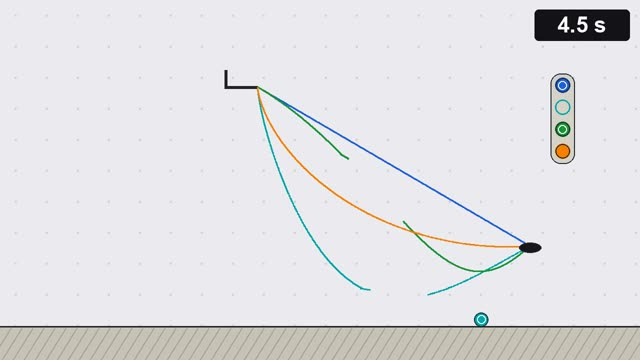} &
\parbox[c]{\linewidth}{\scriptsize\raggedright \textbf{L2:} \emph{Which of these balls jumps across a gap and lands on the far side?} GT: the green ball.\\[1pt]
\textbf{L5:} \emph{If every ball started with a push three times as strong, how many balls would fall through a gap?} GT: 1.}\\[2pt]
\rotatebox{90}{\parbox{0.13\linewidth}{\centering\scriptsize Stacking collapse}} &
\includegraphics[width=\linewidth]{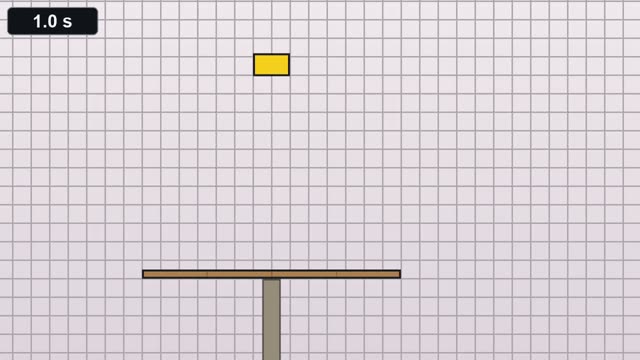} &
\includegraphics[width=\linewidth]{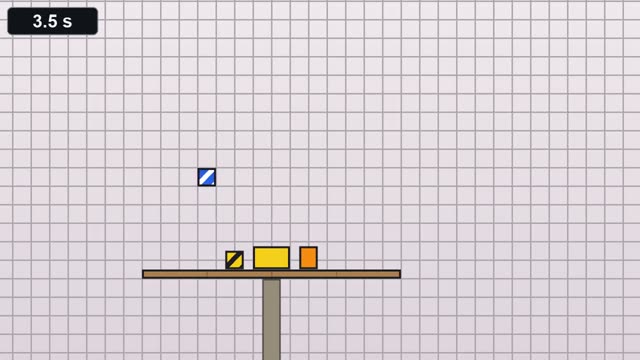} &
\includegraphics[width=\linewidth]{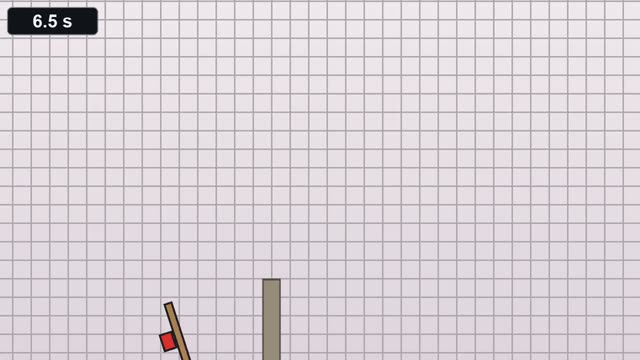} &
\parbox[c]{\linewidth}{\scriptsize\raggedright \textbf{L2:} \emph{Which way does the plank tip?} GT: to the left.\\[1pt]
\textbf{L5:} \emph{Suppose the plain red block had been dropped one grid square further to the right. Would the plank tip over right after it lands?} GT: no.}\\[2pt]
\rotatebox{90}{\parbox{0.13\linewidth}{\centering\scriptsize Spinning solids}} &
\includegraphics[width=\linewidth]{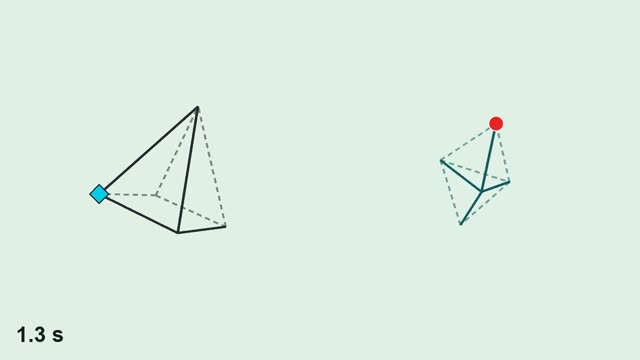} &
\includegraphics[width=\linewidth]{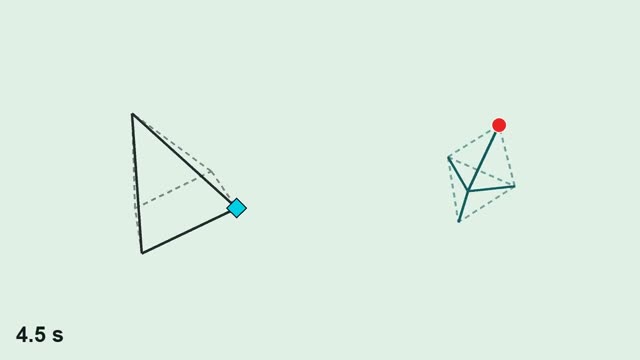} &
\includegraphics[width=\linewidth]{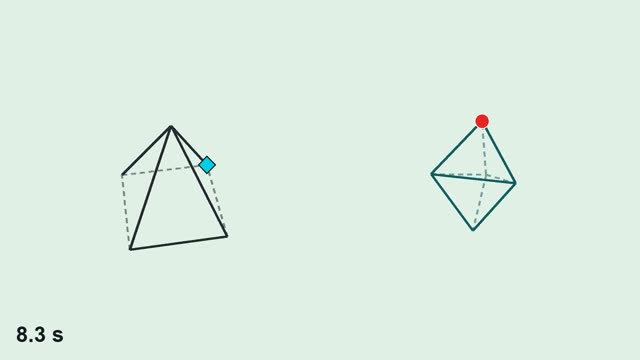} &
\parbox[c]{\linewidth}{\scriptsize\raggedright \textbf{L2:} \emph{Which way do the two solids spin?} GT: the left solid's near side moves left, the right solid's moves right.\\[1pt]
\textbf{L3:} \emph{How many complete turns does the solid on the right make during the clip?} GT: 2.}\\[2pt]
\end{tabular}
\caption{\textbf{Example clips: mechanics, fluids, and 3D geometry.} Three frames per clip and two of its ten questions (GT: ground truth). Pour order, a crack, the drain, a jump over a gap, the tipping direction, and the spin direction all have to be seen as they happen.}
\label{fig:examples_a}
\end{figure*}

\begin{figure*}[p]
\centering
\setlength{\tabcolsep}{1.5pt}
\renewcommand{\arraystretch}{0.6}
\begin{tabular}{@{}c *{3}{>{\centering\arraybackslash}m{0.215\linewidth}} >{\raggedright\arraybackslash}m{0.28\linewidth}@{}}
& {\scriptsize $t=0.15\,T$} & {\scriptsize $t=0.50\,T$} & {\scriptsize $t=0.92\,T$} & {\scriptsize Questions from the clip's ladder}\\[2pt]
\rotatebox{90}{\parbox{0.13\linewidth}{\centering\scriptsize Multi-agent chase}} &
\includegraphics[width=\linewidth]{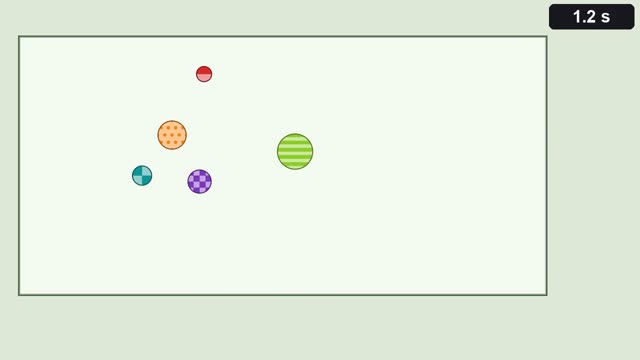} &
\includegraphics[width=\linewidth]{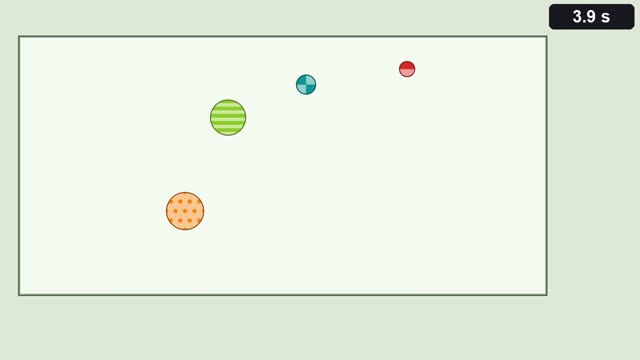} &
\includegraphics[width=\linewidth]{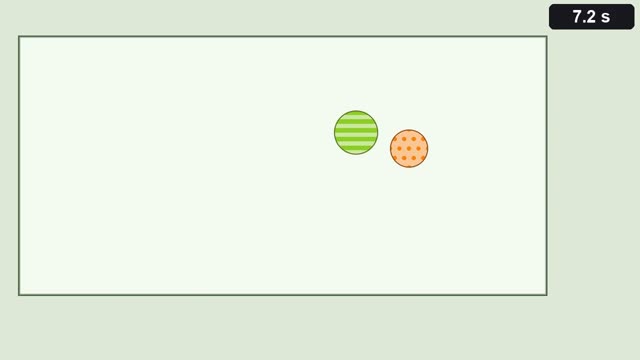} &
\parbox[c]{\linewidth}{\scriptsize\raggedright \textbf{L4:} \emph{In what order are the balls caught?} GT: purple checkered, then red two-tone, then teal quartered.\\[1pt]
\textbf{L5:} \emph{Suppose the lime striped ball moved at half its speed the whole time. How many balls would it catch?} GT: 2.}\\[2pt]
\rotatebox{90}{\parbox{0.13\linewidth}{\centering\scriptsize Tangled braid}} &
\includegraphics[width=\linewidth]{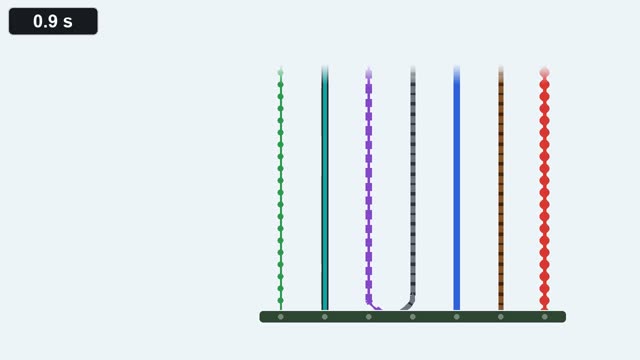} &
\includegraphics[width=\linewidth]{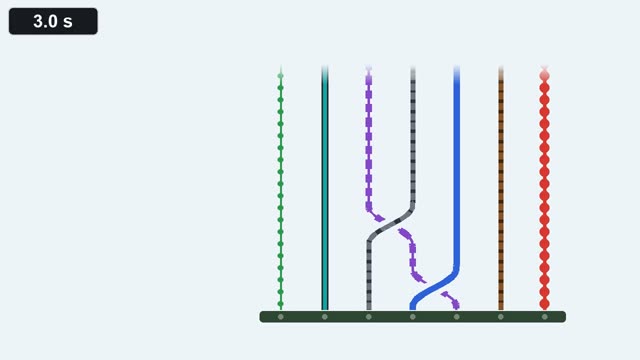} &
\includegraphics[width=\linewidth]{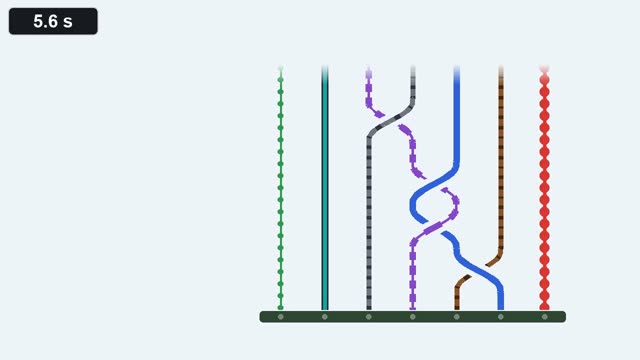} &
\parbox[c]{\linewidth}{\scriptsize\raggedright \textbf{L2:} \emph{Which strand does the blue strand cross first?} GT: purple.\\[1pt]
\textbf{L4:} \emph{Which strand passes over other strands the most times?} GT: blue.}\\[2pt]
\rotatebox{90}{\parbox{0.13\linewidth}{\centering\scriptsize Blinking networks}} &
\includegraphics[width=\linewidth]{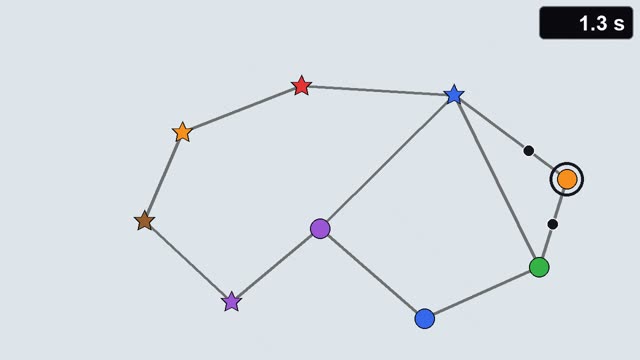} &
\includegraphics[width=\linewidth]{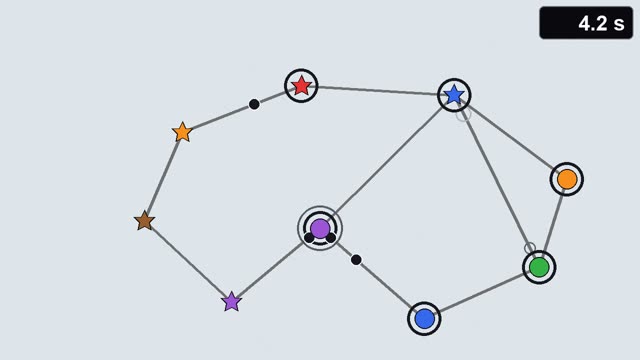} &
\includegraphics[width=\linewidth]{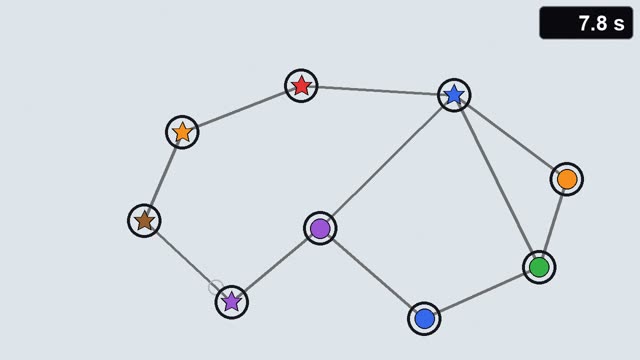} &
\parbox[c]{\linewidth}{\scriptsize\raggedright \textbf{L2:} \emph{Of the nodes directly linked to the red star, which one lights up first?} GT: blue star.\\[1pt]
\textbf{L4:} \emph{Which node sent the pulse that lights up the brown star?} GT: orange star.}\\[2pt]
\rotatebox{90}{\parbox{0.13\linewidth}{\centering\scriptsize Cellular automata}} &
\includegraphics[width=\linewidth]{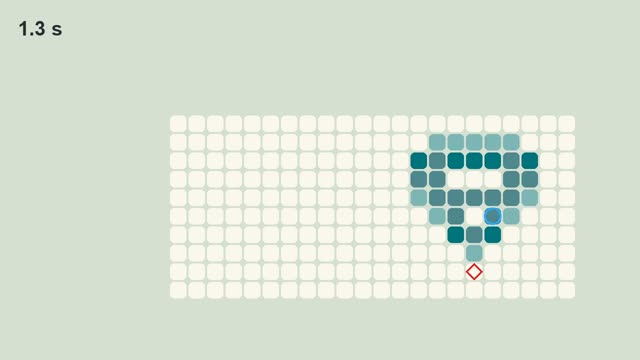} &
\includegraphics[width=\linewidth]{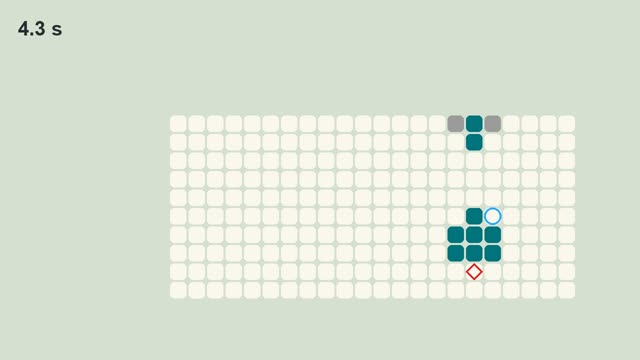} &
\includegraphics[width=\linewidth]{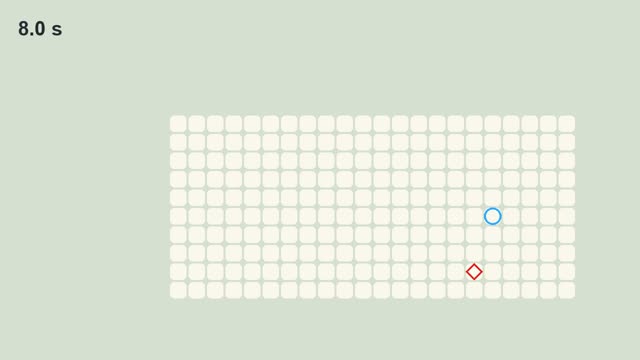} &
\parbox[c]{\linewidth}{\scriptsize\raggedright \textbf{L4:} \emph{At about what time does the grid first become completely empty?} GT: 7.4\,s.\\[1pt]
\textbf{L5:} \emph{Suppose the cell with the red diamond frame had been teal at the start. How many teal cells would there be after the first step?} GT: 19.}\\[2pt]
\rotatebox{90}{\parbox{0.13\linewidth}{\centering\scriptsize Shell game}} &
\includegraphics[width=\linewidth]{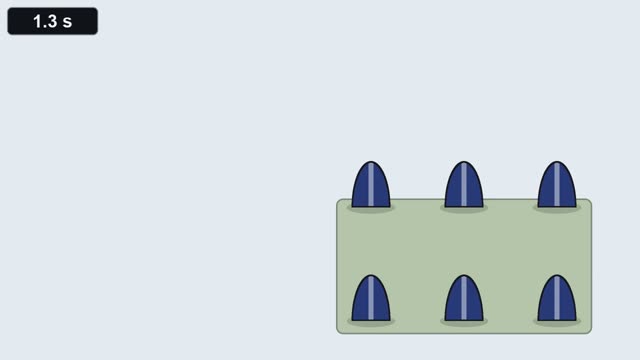} &
\includegraphics[width=\linewidth]{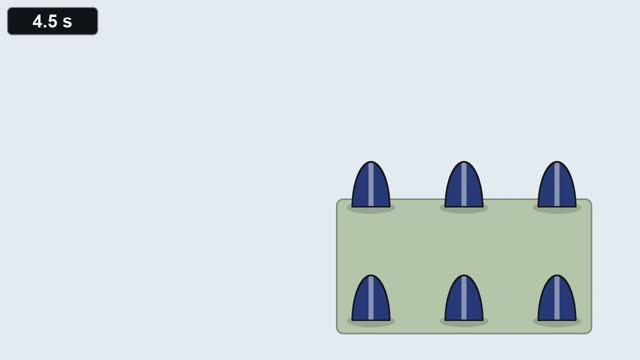} &
\includegraphics[width=\linewidth]{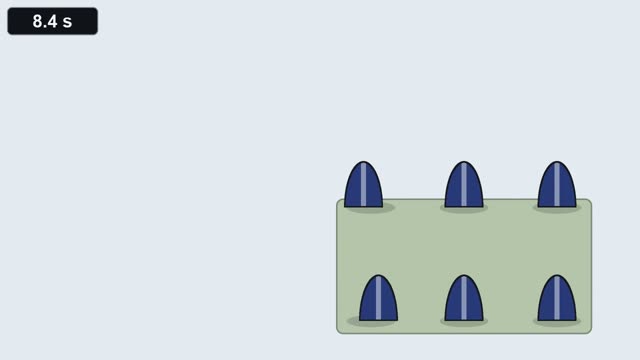} &
\parbox[c]{\linewidth}{\scriptsize\raggedright \textbf{L4:} \emph{Before the last swap, does the cup hiding the orange triangle ever come back to the position it started in?} GT: no.\\[1pt]
\textbf{L5:} \emph{Suppose the fourth and fifth swaps had happened in the opposite order. Under the cup in which position would the triangle be?} GT: the bottom-left position.}\\[2pt]
\rotatebox{90}{\parbox{0.13\linewidth}{\centering\scriptsize Symmetry pattern}} &
\includegraphics[width=\linewidth]{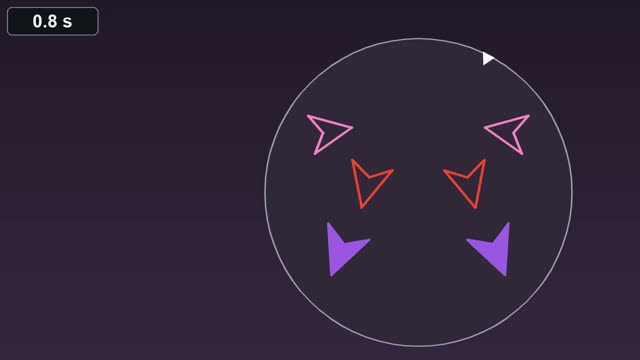} &
\includegraphics[width=\linewidth]{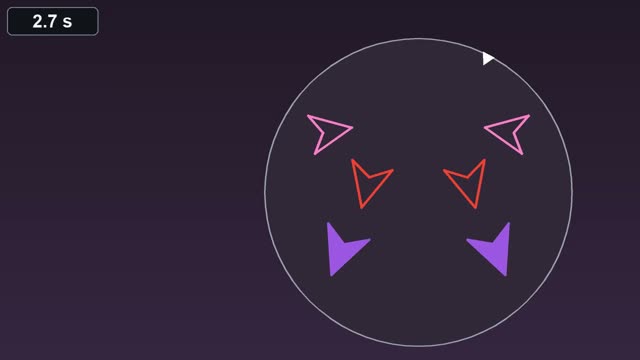} &
\includegraphics[width=\linewidth]{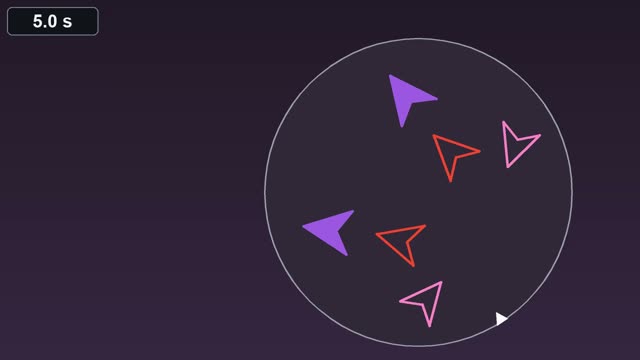} &
\parbox[c]{\linewidth}{\scriptsize\raggedright \textbf{L2:} \emph{What is the first move the disc makes?} GT: a left-right flip.\\[1pt]
\textbf{L4:} \emph{By the end of the disc's third move, how many degrees has it turned in total (flips don't count)?} GT: 120$^\circ$.}\\[2pt]
\end{tabular}
\caption{\textbf{Example clips: agents, rules, propagation, and occlusion.} Catch order, strand crossings, the pulse that lit a node, the step at which the grid empties, cup swaps, and the disc's turns are properties of the whole clip, not of any single frame.}
\label{fig:examples_b}
\end{figure*}

\cref{fig:teaser} in the main paper shows one clip per reasoning family; \cref{fig:examples_a,fig:examples_b} show twelve sub-benchmarks, one clip
each, with two questions from its ladder. Each clip is the first clip (seed~0) of its sub-benchmark, and the frames are
cut at 15\%, 50\%, and 92\% of the clip. The clock in the
corner is the only on-screen text, apart from scale labels where a quantity
must be read, as on the graduated glasses. All 24 questions shown are flagged as requiring motion, and the clips,
questions, and answers are identical to those in the release.

\section{Reasoning families in detail}
\label{sec:families_detail}

\noindent\textbf{Mechanics and motion.} Restitution in bouncing objects,
elastic multi-body collisions that swap fill patterns, gear trains with
reversals, slips, and brakes, springs and pendulums modeled with simple
harmonic motion, wave pulses reflecting and splitting in bead chains,
torque on a balance scale, balls racing down curved tracks with jumpable
gaps, rolling bodies whose acceleration depends on rotational inertia,
and block towers that topple when the modeled center of mass crosses
a support edge.

\noindent\textbf{Fluids and optics.} Buoyancy and displacement in a water
tank, taps and $\sqrt{h}$-dependent drainage in communicating vessels,
pouring from jugs into graduated glasses with measurable liquid levels,
fluid levelling in a flooded depth grid, light pulses reflecting across
visible mirrors, and inferring hidden mirror positions from pulse exit
locations.

\noindent\textbf{Spatial and 3D geometry.} Temporal extensions of GRIP's
image categories: rotating wireframe solids whose structure and spin
direction are inferred across frames, voxel structures that reveal
hidden cubes during turntable rotation, shadows cast by a moving sun
or lamp that merge on the ground, and symmetric patterns that are
constructed, rotated, or flipped, sometimes with one odd shape.

\noindent\textbf{Occlusion and object permanence.} Amodal pattern counting
behind sliding occluders, shapes that change while hidden behind translucent
panels, counting overlapping shapes and determining their stacking order,
shell games, multiple-object tracking with exact elastic bounces, and
following bagel pieces behind a screen.

\noindent\textbf{Temporal events and propagation.} Counting flashes among
decoy pulses, nested-contour rotation and mark patterns, signals spreading
through a network with hidden slow links, chip-firing sandpile cascades, and a
hidden cellular-automaton rule to be induced from its generations.

\noindent\textbf{Agents and navigation.} Telling catchers from prey from motion
alone, the draw order of grid-routed colored routes, traffic-light queues,
rope-swing path planning, and robots hopping toward a foul line under a
stopping rule.

\noindent\textbf{Rule tracking and algorithms.} Simultaneous-move chess
``dances,'' over/under crossings in rope braids, tile shuffles in which lines
form and break along the way, conveyor sorting by a rule that must be induced
while other properties act as distractors, and a running-record archery
contest.

\cref{tab:spotlight} describes six sub-benchmarks in detail. They illustrate
three sources of difficulty that recur across \bench. First, \emph{properties
that require information across frames}: a gear's rotation direction depends
on its position along the mesh, and the amount a jug delivers depends on its
remaining contents after earlier pours. Second, \emph{rules that must be
induced before they are applied}: the cellular-automaton rule is drawn from
up to 24 candidates per variant and is never shown. Third, \emph{events
whose meaning depends on the preceding state}: a chess move is legal only
if its path was free when the round began, and a flooded section may fill
only from a full, higher neighbor. These examples have parallels in
physical-AI settings, including braked transmissions, leaking containers,
changing viewpoints, and simultaneous actions.

\begin{table*}[p]

\caption{\textbf{Six sub-benchmarks in detail.} \emph{Variants}: the
scenario variants, drawn uniformly per clip. \emph{Re-sampled per clip}:
the main randomized parameters. \emph{Q--A}: distinct question--answer pairs
among the $25{,}000$ released questions. \emph{Acc.}: strict accuracy (\%),
mean across eleven models / highest accuracy on that sub-benchmark,
with the constant-answer baseline in parentheses.}
\label{tab:spotlight}
\centering
\scriptsize
\setlength{\tabcolsep}{3pt}
\renewcommand{\arraystretch}{1.15}
\begin{tabular}{@{}>{\raggedright\arraybackslash}p{0.095\linewidth}>{\raggedright\arraybackslash}p{0.2\linewidth}>{\raggedright\arraybackslash}p{0.27\linewidth}>{\raggedright\arraybackslash}p{0.235\linewidth}rr@{}}
\toprule
\textbf{Sub-benchmark} & \textbf{Variants} & \textbf{Re-sampled per clip} & \textbf{What makes it hard} & \textbf{Q--A} & \textbf{Acc.}\\
\midrule
\rowcolor{rowgrey}
\emph{Gear train} (mechanics)
& steady; driver reversal; one gear slides out of mesh and its sub-train spins down; brake on the teeth stops the whole train; motor speed change
& 3--8 involute gears with 10--36 teeth; chain, hub, or tree topology; optional loose gear; driver speed and direction; 9 colors, 4 face patterns, 11 themes $\times$ 7 textures
& Direction alternates along the mesh and speed scales with $1/\text{teeth}$; a disengaged gear stops everything it drives but nothing upstream
& 7{,}162 & 47.7 / 58.8 (27.8)\\
\emph{Glasses fill} (fluids)
& pours into empty glasses; pours stacked as layers; a jug that is empty or runs dry mid-pour; a glass that cracks and drains to the crack
& 4--10 graduated 100\,ml glasses; 3--4 jugs with 8 patterns and 8 liquid colors; 40--150\,ml per jug; 2--4 pours; stream rate 55--90\,ml/s; crack height and drain rate
& A pour delivers $\min(\text{asked},\text{left})$; levels are read off a scale; layered colors never mix; a crack removes only the liquid above it
& 4{,}129 & 53.8 / 82.0 (27.5)\\
\rowcolor{rowgrey}
\emph{Well flood} (fluids)
& one entry; a second, delayed entry; the pour stops or doubles; walls block spills; one physically impossible flow event (violation)
& $5{\times}5$ to $8{\times}8$ depth grid with random, cone, or tilted terrain; entry section, start time, and pour rate; 2--5 walls; violation type, section, and time
& Water fills sections in order of depth and spills only to lower neighbors; the violation is not marked when it happens
& 7{,}942 & 42.6 / 60.2 (27.1)\\
\emph{Spinning solids} (3D geometry)
& constant spin; stop and reverse; speed-up or slow-down; two solids side by side; switch of rotation axis
& 21 wireframe solids; size, position, tilt, and phase; camera elevation 6--28$^\circ$; three depth cues, including none (depth only from motion); 1--3 corner markers
& Spin direction, turn count, and changes in rotation speed or axis are inferred across frames; structure may also be ambiguous from a single view
& 2{,}936 & 39.7 / 48.0 (27.0)\\
\rowcolor{rowgrey}
\emph{Cellular automata} (temporal)
& Life-like; wrap-around torus; 4-neighbor; three-state ``generations''; 1-D elementary rules drawn as a space-time diagram
& rule from a pool of 8--24 per variant (67 variant--rule pairs); $9{\times}11$ to $15{\times}41$ grid; 6--12 generations; start layout and density; 2--3 marked cells
& The rule is never shown: it must be induced from the generations and then applied, including to counterfactual start states
& 16{,}665 & 40.3 / 50.9 (27.7)\\
\emph{Chess dance} (rules)
& every move legal; one illegal move (violation); one mover per round; captures; holes cut into the board
& $6{\times}6$ to $8{\times}8$ board; 3--9 pieces of five types, each a unique color; 1--3 simultaneous movers per round; 4--9 rounds; board style and piece style
& Several pieces move at once, and legality depends on which squares were occupied when the round began
& 10{,}985 & 44.2 / 63.0 (26.7)\\
\bottomrule
\end{tabular}
\end{table*}

\begin{table}[h]
\caption{\textbf{Questions per ladder rung} in the full release
(1{,}000{,}000 questions) and the share flagged as requiring motion.}
\label{tab:levels}
\centering
\footnotesize
\setlength{\tabcolsep}{4pt}
\begin{tabular}{@{}lrrrrr@{}}
\toprule
& \textbf{L1} & \textbf{L2} & \textbf{L3} & \textbf{L4} & \textbf{L5}\\
\midrule
\rowcolor{rowgrey}
Questions & 174{,}317 & 241{,}383 & 240{,}842 & 171{,}720 & 171{,}738\\
Share (\%) & 17.4 & 24.1 & 24.1 & 17.2 & 17.2\\
\rowcolor{rowgrey}
Motion (\%) & 33.0 & 91.0 & 92.1 & 94.6 & 87.4\\
\bottomrule
\end{tabular}
\end{table}

\section{Release scale}
\label{sec:scale}

Every sub-benchmark has $2{,}500$ clips (seeds 0--2{,}499) with ten questions
each, for $100{,}000$ clips and one million questions in total. Clips last
7.7\,s on average (per sub-benchmark means range from 5.4 to 9.5\,s), for
about 213 hours and 15.8\,GB of H.264 video (CRF~23). Rendering is CPU-only: a clip takes 5.8\,s on average to draw and
encode, and one sub-benchmark takes a median of 46 minutes on a 4-CPU job
split into four parallel shards. Rendering video still costs far more per
sample than rendering a static image, so the release is somewhat below
GRIP's scale of about $2{,}900$ images per domain. 
The release size is fixed at $2{,}500$ clips per sub-benchmark, while
seeded generation allows additional samples to be produced. Each sub-benchmark's validator ran over all $2{,}500$ clips
and $25{,}000$ questions of the release and found zero mismatches.

\section{Sub-benchmark catalogue}
\label{sec:catalogue}

\cref{tab:catalogue} lists all 40 \bench sub-benchmarks and whether they
include injected single-violation clips. Every sub-benchmark has $2{,}500$
clips with ten questions each ($25{,}000$ questions), spread over all five
ladder levels, and every row passes its validator with zero mismatches. Six
sub-benchmarks adapt Jane Street puzzles:
\emph{tangled braid}, \emph{hidden mirrors}, \emph{rainbow bagel},
\emph{swing time}, \emph{robot long jump}, and \emph{robot archery}.

\begin{table*}[p]
\caption{\textbf{All 40 \bench sub-benchmarks.} Each has $2{,}500$ clips and
$25{,}000$ questions on the five-rung ladder. \emph{V}: injected
single-violation clips present.}
\label{tab:catalogue}
\centering
\scriptsize
\setlength{\tabcolsep}{3pt}
\renewcommand{\arraystretch}{1.0}
\begin{tabular}{@{}llcp{0.52\linewidth}@{}}
\toprule
\textbf{Family} & \textbf{Sub-benchmark} & \textbf{V} & \textbf{What must be tracked over time}\\
\midrule
\multirow{9}{*}{Mechanics \& motion}
 & bouncing ball restitution & -- & Bounce counts and energy loss under drops, clamped releases, colored pads, and shelves\\
 & shape collision & -- & Elastic multi-body collisions; colliding shapes swap fill patterns\\
 & gear train & -- & Direction and speed through 3--8 meshed gears; reversals, slips, and brakes\\
 & simple harmonic motion & -- & 3--6 springs and pendulums, damped or staggered; periods, amplitudes, crossings\\
 & spring chain waves & -- & Pulse reflection, inversion, and splitting where bead thickness changes\\
 & balance scale weighing & -- & Torque as dot-weighted shapes are dropped, moved, or removed; solving a hidden weight\\
 & rolling races & -- & Races on 2--10 frictionless curved tracks with jumpable gaps\\
 & inertia race & -- & Blocks, balls, discs, shells, and rings racing down ramps; $a=g\sin\theta/(1+k)$\\
 & stacking collapse & -- & Blocks stacked or pulled out until the center of mass leaves its support\\
\midrule
\multirow{6}{*}{Fluids \& optics}
 & buoyancy tank & -- & Objects of hidden density dropped, released, filled, or drained; float order and heights\\
 & vessels and clocks & -- & Taps, steady and $\sqrt{h}$ drains, and pipe-joined vessels that equalize\\
 & glasses fill & -- & Jugs poured into graduated glasses; reading the scale, layers, dry and cracked glasses\\
 & well flood trace & \cmark & Flood-fill order by depth; premature-fill or uphill-flow violation\\
 & mirror beam reflection & -- & Light pulse on a mirror board with splitters, one-sided and pivoting mirrors\\
 & hidden mirrors & -- & Inferring invisible mirrors from where fired pulses turn and exit\\
\midrule
\multirow{4}{*}{Spatial \& 3D geometry}
 & spinning polyhedra & -- & Faces, edges, and spin direction of 21 wireframe solids that stop, reverse, or switch axis\\
 & turntable voxels & -- & Hidden and visible cubes of a stack revealed by a turntable\\
 & moving shadows & -- & Matching merged shadows to objects under a moving sun or lamp\\
 & symmetry pattern video & \cmark & Symmetric pictures built, turned, or flipped; one odd shape in about 45\% of clips\\
\midrule
\multirow{6}{*}{Occlusion \& permanence}
 & occluded pattern counting & -- & Amodal counting of 7--42 shapes behind sliding occluders\\
 & translucent occlusion & -- & Shapes that pass, stay, or change color, shape, or size behind translucent panels\\
 & overlapping shapes & -- & Counting, stacking order, and arrival or departure order of overlapping shapes\\
 & shell game tracking & -- & Tokens hidden under 3--6 identical cups through swaps, turns, and peeks\\
 & object tracking mot & -- & Tracking objects through elastic bounces, screens, and orbits\\
 & rainbow bagel & -- & Following bagel pieces behind a screen; ring, quarter, and exact areas\\
\midrule
\multirow{5}{*}{Temporal events \& propagation}
 & temporal flash counting & -- & Flash counts among decoy pulses; relays with hidden trigger chains\\
 & nested shapes & -- & Nested-outline counting, marks placed and removed, rotation patterns\\
 & blinking networks & -- & A flash spreading on a graph with slow, one-way, and hidden links\\
 & sandpile cascade & -- & Chip-firing cascades as grains are dropped one at a time\\
 & cellular automata & -- & Inducing a hidden rule from the generations shown, in five neighborhood types\\
\midrule
\multirow{5}{*}{Agents \& navigation}
 & multi agent chase & -- & Catchers and prey from motion alone; who caught whom, when, and in what order\\
 & colored route puzzle & -- & Draw order of grid-routed colored routes, some erased later\\
 & traffic light queue & -- & Vehicles queueing at lights, crossroads, and lights in a row\\
 & swing time & -- & Rope-swing route across posts; rope lengths and pivot order\\
 & robot long jump & -- & Random hops, a hidden threshold, a foul line, and scores\\
\midrule
\multirow{5}{*}{Rule tracking \& algorithms}
 & chess dance trace & \cmark & 3--9 pieces moving at once; one illegal move in about 20\% of clips\\
 & tangled braid & -- & Over/under crossings of 4--8 strands as the braid scrolls away\\
 & tile alignment shuffle & -- & Tile swaps, rotations, and slides with false tiles; lines that form and break\\
 & conveyor belt sort & -- & Inducing a sorting rule from objects routed into trays\\
 & robot archery & -- & Running-record elimination in queue order\\
\midrule
\textbf{Total} & 40 sub-benchmarks & \textbf{3} & $100{,}000$ clips, $1{,}000{,}000$ questions\\
\bottomrule
\end{tabular}
\end{table*}

\section{Scenario variants of every sub-benchmark}
\label{sec:variants_full}

\cref{tab:complexity} lists the scenario variants of all 40 sub-benchmarks, the number of distinct
question--answer pairs among its $25{,}000$ released questions, the share of
its questions flagged as requiring motion, and the mean strict accuracy
of the eleven models. Question diversity is not what makes a
sub-benchmark hard: \emph{rainbow bagel} has the fewest distinct
question--answer pairs and is among the three hardest sub-benchmarks, while
\emph{cellular automata} has the most and lies near the middle of the
difficulty range (40.3\%, against 32.9--55.3\% over all sub-benchmarks).

\begin{table*}[p]
\caption{\textbf{Scenario variants and question diversity of every
sub-benchmark.} \emph{Var.}: number of scenario variants. \emph{Q--A}:
distinct question--answer pairs among the $25{,}000$ released questions.
\emph{Motion}: share of questions flagged as requiring motion (\%).
\emph{Acc.}: mean strict accuracy of the eleven models (\%).}
\label{tab:complexity}
\centering
\scriptsize
\setlength{\tabcolsep}{4pt}
\begin{tabular}{@{}lr>{\raggedright\arraybackslash}p{0.5\linewidth}rrr@{}}
\toprule
\textbf{Sub-benchmark} & \textbf{Var.} & \textbf{Scenario variants} & \textbf{Q--A} & \textbf{Motion} & \textbf{Acc.}\\
\midrule
\multicolumn{6}{@{}l}{\emph{Mechanics \& motion}}\\
\rowcolor{rowgrey}
bouncing ball restitution & 4 & drop, staggered, pads, throw & 15{,}238 & 90.8 & 45.0\\
shape collision & 5 & swap, infect, posts, holes, rails & 14{,}620 & 86.9 & 39.4\\
\rowcolor{rowgrey}
gear train & 5 & steady, reversal, disengage, brake, speed change & 7{,}162 & 77.7 & 47.7\\
simple harmonic motion & 5 & horizontal, vertical, pendulum, damped, staggered & 13{,}752 & 84.5 & 38.1\\
\rowcolor{rowgrey}
spring chain waves & 4 & reflect, junction, collide, train & 2{,}232 & 95.9 & 42.0\\
balance scale weighing & 5 & drop, add remove, transfer, lever, mystery & 9{,}111 & 73.6 & 54.4\\
\rowcolor{rowgrey}
rolling races & 5 & plain, jumps, double, all gaps, steep & 5{,}357 & 79.8 & 38.1\\
inertia race & 5 & fair start, handicap, stagger, pursuit, twin & 9{,}906 & 82.9 & 45.0\\
\rowcolor{rowgrey}
stacking collapse & 4 & tower, twin, pull out, plank & 9{,}251 & 87.8 & 51.9\\
\midrule
\multicolumn{6}{@{}l}{\emph{Fluids \& optics}}\\
\rowcolor{rowgrey}
buoyancy tank & 4 & drop, rise, fill, drain & 13{,}479 & 88.3 & 50.4\\
vessels and clocks & 5 & taps drains, drain race, fill race, joined pair, two pairs & 7{,}782 & 91.5 & 48.0\\
\rowcolor{rowgrey}
glasses fill & 4 & row pours, layer stack, dry jug, cracked glass & 4{,}129 & 70.1 & 53.8\\
well flood trace & 5 & valid flow, two entries, pour change, walled, broken flow & 7{,}942 & 82.6 & 42.6\\
\rowcolor{rowgrey}
mirror beam reflection & 5 & classic, one sided, splitter, flip, two beams & 8{,}967 & 54.4 & 40.9\\
hidden mirrors & 5 & classic, rect, volley, blocks, partial & 8{,}375 & 94.7 & 41.8\\
\midrule
\multicolumn{6}{@{}l}{\emph{Spatial \& 3D geometry}}\\
\rowcolor{rowgrey}
spinning polyhedra & 5 & steady, reverse, speed change, pair, axis switch & 2{,}936 & 69.1 & 39.7\\
turntable voxels & 5 & spin, stops, reverse, build, remove & 2{,}478 & 73.7 & 41.0\\
\rowcolor{rowgrey}
moving shadows & 5 & floating, posts, drifting, rising, lamp & 8{,}141 & 97.4 & 32.9\\
symmetry pattern video & 4 & waves, reveal, assemble, turns & 4{,}811 & 82.3 & 47.2\\
\midrule
\multicolumn{6}{@{}l}{\emph{Occlusion \& permanence}}\\
\rowcolor{rowgrey}
occluded pattern counting & 4 & slide, two occluders, gaps, odd one & 4{,}055 & 79.6 & 46.0\\
translucent occlusion & 4 & crossing, hidden change, hide and stay, two panels & 11{,}074 & 83.6 & 53.5\\
\rowcolor{rowgrey}
overlapping shapes & 4 & arrive, paper, depart, glide & 12{,}476 & 68.9 & 55.3\\
shell game tracking & 5 & classic, multi, rotate, grid, peek & 9{,}492 & 90.0 & 42.0\\
\rowcolor{rowgrey}
object tracking mot & 4 & colours, targets, occluders, orbit & 8{,}390 & 81.0 & 53.2\\
rainbow bagel & 4 & wide hole, one by one, bunches, long hide & 1{,}007 & 55.1 & 35.0\\
\midrule
\multicolumn{6}{@{}l}{\emph{Temporal events}}\\
\rowcolor{rowgrey}
temporal flash counting & 5 & grid, scatter, relay, signal, drift & 12{,}241 & 94.1 & 41.5\\
nested shapes & 5 & assemble, spin, marks, recolor, drift & 9{,}829 & 83.4 & 42.1\\
\rowcolor{rowgrey}
blinking networks & 5 & hidden slow, marked speeds, one way, two sources, double trigger & 15{,}489 & 89.4 & 43.9\\
sandpile cascade & 4 & row, grid, ring, pyramid & 8{,}969 & 86.5 & 39.0\\
\rowcolor{rowgrey}
cellular automata & 5 & life like, torus, von neumann, generations, elementary & 16{,}665 & 78.1 & 40.3\\
\midrule
\multicolumn{6}{@{}l}{\emph{Agents \& navigation}}\\
\rowcolor{rowgrey}
multi agent chase & 4 & open, pillars, safe zone, bigger & 12{,}255 & 89.1 & 45.6\\
colored route puzzle & 5 & sequential, overlap, dead ends, erase, traveller & 5{,}124 & 72.1 & 48.2\\
\rowcolor{rowgrey}
traffic light queue & 5 & single, two lanes, crossing, mixed, two lights & 5{,}452 & 82.8 & 52.6\\
swing time & 5 & left right, bottom top, corner, dense, pits & 6{,}099 & 78.8 & 38.3\\
\rowcolor{rowgrey}
robot long jump & 4 & turns, together, staggered, steady & 3{,}003 & 44.9 & 39.5\\
\midrule
\multicolumn{6}{@{}l}{\emph{Rules \& algorithms}}\\
\rowcolor{rowgrey}
chess dance trace & 5 & dance, rogue, relay, capture, holes & 10{,}985 & 82.4 & 44.2\\
tangled braid & 5 & classic, parallel, sideways, pauses, pure & 10{,}660 & 91.4 & 43.4\\
\rowcolor{rowgrey}
tile alignment shuffle & 5 & swap, neighbour, cycle, slide, double & 10{,}415 & 84.6 & 33.9\\
conveyor belt sort & 5 & size, colour, shape, three way, alternate & 14{,}179 & 78.5 & 43.5\\
\rowcolor{rowgrey}
robot archery & 4 & classic, two lives, stay on, big field & 4{,}610 & 87.2 & 41.3\\
 
\midrule
\textbf{Total} & \textbf{185} & & \textbf{348{,}138} & \textbf{81.1} & \\
\bottomrule
\end{tabular}
\end{table*}

\section{Model access and configuration}
\label{sec:access}

\paragraph{Evaluation subset.}
Every model answers the same fixed subset of the release: 250 clips sampled
from each sub-benchmark's $2{,}500$ released clips with a fixed
random seed, yielding $10{,}000$ clips with ten
questions each. This gives $100{,}000$ questions per model: $17{,}394$ on
L1, $24{,}150$ on L2, $24{,}141$ on L3, $17{,}244$ on L4, and $17{,}071$
on L5. Of these, 96.9\% have four options, 0.8\% three, and 2.3\% two.
The motion flag applies to 81.1\% of the subset,
matching the full release to one decimal place.

\cref{tab:access} lists the served build of each model. The Aleph models
(including Molmo2-8B) run
on Aleph, a hosted inference service. The NRP models run
on the managed-LLM service of the National Research Platform. Both expose an OpenAI-compatible
API and do not report a build hash, so the
upstream names are those listed by the services. Model details are taken from
the services' model pages, and the serving
hardware is not disclosed. A
further model, Kimi-K2.7-Code on NRP, was excluded because every request with a video returned an HTTP~500
error while text-only requests worked.

The other four models run on our own GPU cluster, each behind a vLLM
OpenAI-compatible server on the same node, with the same client and
protocol. Nemotron-3.5-Super-VL-120B-A12B (NVFP4) uses two H100 80\,GB GPUs
with tensor parallelism and the model card's recipe, including multi-token
speculative decoding and a video-token pruning rate of 0.5. Keye-VL-1.5-8B,
InternVideo2.5-Chat-8B, and Reka Edge 2603 each use one H100. Nemotron and
Keye read the clip at 2\,fps (at most 256 frames), Reka Edge reads 16
frames, and InternVideo2.5 reads 32. Thinking is turned off with a request
flag (Nemotron) or a prompt suffix (Keye). vLLM supports video for
InternVL-style models only with a Qwen language model, so we serve
InternVideo2.5 (an InternLM2 model) through a small vLLM plugin that enables
video and ports its HiCo token merging (each group of four frames merged
from 1{,}024 to 64 vision tokens). The plugin is not bit-exact: on 50
questions it agrees with the original Transformers implementation on 78\%
of answers. Its merged features match the original's (cosine 0.988) as
closely as two runs of the original match each other, because the merge
depends on near-tied bf16 scores. Reka Edge
is served with vLLM~0.17.1 and the vendor's plugin, the others with
vLLM~0.32.0.

\begin{table}[t]
\caption{\textbf{Served model builds.} Context and output limits are those
listed by the service, or the server limit we set (self-hosted models).}
\label{tab:access}
\centering
\scriptsize
\setlength{\tabcolsep}{2pt}
\begin{tabular}{@{}lll>{\raggedright\arraybackslash}p{0.27\linewidth}@{}}
\toprule
\textbf{Model} & \textbf{Weights} & \textbf{Ctx. / out.} & \textbf{Architecture}\\
\midrule
\rowcolor{rowgrey}
Qwen3.6-27B & BF16 & 131{,}072 / 32{,}768 & 27B dense, Gated-DeltaNet hybrid\\
Qwen3.8-27B & FP8 & 524{,}288 / 131{,}072 & 27B dense, 64 layers (16$\times$ Gated DeltaNet + 1$\times$ gated attention)\\
\rowcolor{rowgrey}
Gemma-4-26B-A4B & FP8 & 131{,}072 / 8{,}192 & MoE, 25.2B total / 3.8B active, instruction-tuned\\
Qwen3.8-Flash-Next & FP8 & 1M / -- & MoE, 180B total / 6B active\\
\rowcolor{rowgrey}
Gemma-4-31B & W4A16 QAT & 262{,}144 / -- & 31B, instruction-tuned\\
Gemma-4-12B & W4A16 QAT & 262{,}144 / -- & 12B, instruction-tuned\\
\rowcolor{rowgrey}
Molmo2-8B & FP32 (ckpt.) & -- & 8B, served by Aleph; served precision not reported\\
Nemotron-3.5-Super-VL & NVFP4 & -- & MoE, 120B total / 12B active, Mamba hybrid\\
\rowcolor{rowgrey}
Keye-VL-1.5-8B & BF16 & 32{,}768 / -- & 8B\\
InternVideo2.5-8B & BF16 & -- & 8B, InternLM2, HiCo token merging\\
\rowcolor{rowgrey}
Reka-Edge-2603 & BF16 & 16{,}384 / -- & --\\
\bottomrule
\end{tabular}
\end{table}

\section{Motion gain and violation metrics}
\label{sec:metric_details}

\paragraph{Motion gain.}
We define the \emph{motion gain} of a model as
\begin{equation}
\Delta_{\mathrm{motion}} = a_{\mathrm{adj}}^{\text{video}} - a_{\mathrm{adj}}^{\text{final frame}},
\label{eq:motiongain}
\end{equation}
and the \emph{order gain} $\Delta_{\mathrm{order}}$ analogously against
the shuffled condition. A positive $\Delta_{\mathrm{motion}}$ indicates
an advantage from video over the held final frame, while a positive
$\Delta_{\mathrm{order}}$ indicates an advantage over shuffled frames.
A value near zero indicates no measured advantage under the
corresponding control, but does not establish that the model ignores
temporal information.

\paragraph{Shuffled-frame control.}
\cref{tab:shuffled} reports the order gain for the three NRP models on the
10{,}000 questions of the final-frame subset. Each model receives the same 16
frames twice, in temporal order and in a fixed random permutation per clip.
With ordered frames, adjusted accuracy stays within 1.4 points of the
full-video run, so 16 frames carry the information the models use. Shuffling
removes 10.6 and 10.1 points for Qwen3.8-Flash-Next and Gemma-4-31B, and
the gain is larger on motion-flagged questions (12.5 and 13.2) than on the
others (9.4 and 7.3). Gemma-4-12B, at 15.4 points with video, loses 1.5.
Excluding questions that refer to the on-screen clock, whose readings a
shuffled clip still shows, leaves the gain at 9.9, 8.9, and 1.8 points.
Lenient scoring changes the gain by at most 0.3 points.

\begin{table}[t]
\caption{\textbf{Shuffled-frame control} (baseline-adjusted accuracy,
points; strict; 25 clips per sub-benchmark, 10{,}000 questions per model).
\emph{Gain}: ordered minus shuffled 16 frames, with a 95\% clip-bootstrap
interval; \emph{Flag.}/\emph{Unfl.}: gain with and without the motion flag;
\emph{No clock}: gain without questions that refer to the on-screen clock.}
\label{tab:shuffled}
\centering
\scriptsize
\setlength{\tabcolsep}{2.1pt}
\begin{tabular}{@{}lrrrcrrr@{}}
\toprule
\textbf{Model} & \textbf{Video} & \textbf{Ord.} & \textbf{Shuf.} & \textbf{Gain [95\% CI]} & \textbf{Flag.} & \textbf{Unfl.} & \textbf{No clock}\\
\midrule
\rowcolor{rowgrey}
Qwen3.8-Flash-Next & 29.4 & 28.0 & 17.5 & 10.6 [9.4, 11.8] & 12.5 & 9.4 & 9.9\\
Gemma-4-31B & 22.9 & 23.1 & 13.0 & 10.1 [9.0, 11.2] & 13.2 & 7.3 & 8.9\\
\rowcolor{rowgrey}
Gemma-4-12B & 15.4 & 14.7 & 13.2 & 1.5 [0.6, 2.5] & 2.8 & 1.8 & 1.8\\
 
\bottomrule
\end{tabular}
\end{table}

\paragraph{Violation questions.}
In the three injected-violation sub-benchmarks, 301 evaluated questions ask
whether a rule was broken and, if so, by which piece, shape, or section.
Each offers exactly one ``nothing went wrong'' option; 97 questions come
from clips containing a violation and 204 from violation-free clips.
We report detection F1, treating any predicted option other than
``nothing went wrong'' as a positive prediction. We also report
localization accuracy, the fraction of violation-positive questions
for which the model identifies the correct culprit. The two metrics
distinguish detecting a rule violation from correctly localizing it.

\section{Classic paradigms and puzzles}
\label{sec:paradigms}

Several \bench sub-benchmarks adapt established paradigms into
procedurally generated videos, including multiple-object
tracking, shell games, chip-firing sandpiles,
and life-like cellular automata. Six others adapt published
Jane Street puzzles, including rope braids,
hidden mirrors, and robot archery. These adaptations preserve the
underlying rules while introducing dynamic scenes and questions
about evolving states and events.

\section{Prompts and scoring}
\label{sec:prompts}

Each request has one user message with two content parts: the clip, as a
base64-encoded MP4 file, and the
text below, with the question and options exactly as stored in the release.
There is no system prompt.
\begin{quote}\small\ttfamily
Which node is the last one to light up?\\
A. blue star\\
B. brown star\\
C. purple star\\
D. orange star\\
Answer with the letter of the correct option only.
\end{quote}
Decoding uses temperature~0 and at most 16 output tokens, with thinking
turned off (\cref{tab:conditions}). On NRP, each run uses a fresh
prompt-cache key, so no cached prompt is shared between runs. Each
model sees each question once at temperature~0, so the confidence intervals
cover sampling of clips, not run-to-run variation. The \emph{strict} scorer accepts a
reply that starts with an option letter A--J, optionally in parentheses and
not followed by another letter. The \emph{lenient} scorer also accepts
``answer is X'' or ``answer: X'', and a letter after leading punctuation, but
not the article ``A'' in running text. Unparsed replies are wrong. The constant-answer baseline of
a (sub-benchmark, rung) cell is the share of its questions whose correct
letter is the most common one in that cell.

\section{Detailed results}
\label{sec:detailed}

\begin{figure}[h]
\centering
\includegraphics[width=\linewidth]{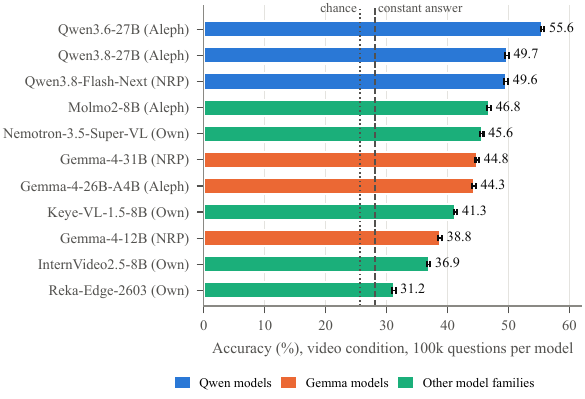}
\caption{\textbf{Overall strict accuracy with 95\% clip-bootstrap
confidence intervals.} Dotted line: chance baseline (25.6\%);
dashed line: constant-answer baseline (28.1\%). Models are
ordered by accuracy.}
\label{fig:overall}
\end{figure}

\begin{figure*}[p]
\centering
\includegraphics[width=\linewidth]{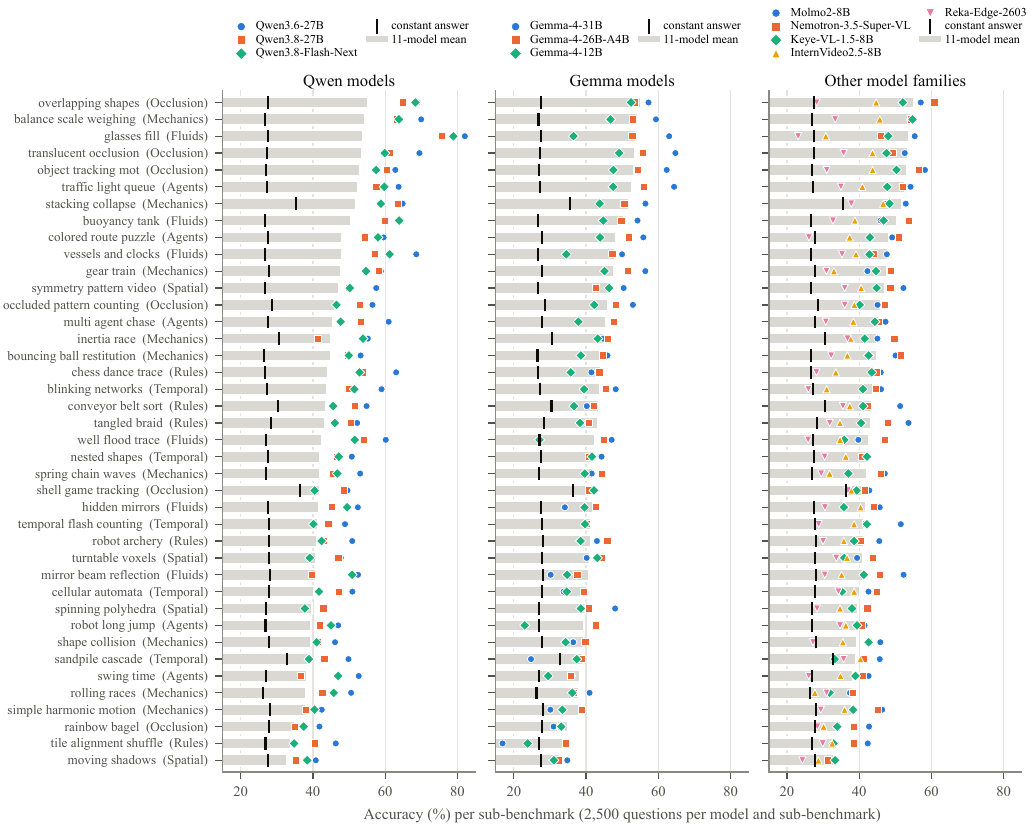}
\caption{\textbf{Accuracy on every sub-benchmark}, sorted by the
eleven-model mean (grey bar, the same in every panel). Markers: individual
models, one panel per group of model families. Black tick: the
constant-answer baseline. Family in parentheses.}
\label{fig:categories}
\end{figure*}

\cref{fig:overall} plots overall accuracy, \cref{tab:ladder,tab:families_full,tab:motion,tab:violation} give the accuracy by rung, by family, by motion flag, and on the violation questions, and \cref{tab:percat} and \cref{fig:categories} give the
accuracy of every model on every sub-benchmark. All values are computed from the released result files.

\paragraph{The ladder is not strictly monotonic.}
Smaller L1--L5 drops for Nemotron (17.6 points)
and Reka Edge (4.6 points), compared with 34--37 points for the Qwen
models, reflect their lower L1 accuracy. The decline is not strictly
monotonic: L3 (measurement) is harder than L4 (multi-step integration)
for the three Qwen models, Nemotron, and Reka, while the two levels
differ by at most 2.0 points for the remaining models. This suggests
that measurement can be as challenging as multi-step integration,
consistent with the image-based GRIP findings.

\paragraph{Sub-benchmarks span a 22-point accuracy range.}
Averaged over the eleven models, the highest-scoring sub-benchmarks are
\emph{overlapping shapes} (55.3\%), \emph{balance scale weighing}
(54.4\%), and \emph{glasses fill} (53.8\%). The lowest-scoring are
\emph{moving shadows} (32.9\%), \emph{tile alignment shuffle} (33.9\%),
and \emph{rainbow bagel} (35.0\%) (\cref{fig:categories}). Qwen3.6-27B leads on 34 of the 40 sub-benchmarks,
Molmo2-8B on four (\emph{rainbow bagel}, \emph{simple harmonic motion},
\emph{tangled braid}, and \emph{temporal flash counting}), and Gemma-4-31B
on two (\emph{spinning polyhedra} and \emph{traffic light queue}).
Gemma-4-31B and Gemma-4-12B fall below the constant-answer baseline on two
sub-benchmarks each, including \emph{tile alignment shuffle} (16.9\% and
23.8\%), and Reka-Edge-2603 on seven.

\paragraph{Format, serving, and frame budget.}
Unparsed responses range from 1.2\% (Qwen3.6-27B) to 15.7\%
(Gemma-4-31B) among Qwen and Gemma models, while the remaining five
models produce at most 28 unparsed replies each. Lenient scoring
recovers at most 0.9 percentage points. On parsed replies alone,
Gemma-4-31B scores 53.2\%, compared with 52.8\% and 52.3\% for the two
Qwen3.8 variants, although these scores exclude different subsets of
questions. The Qwen3.8 variants achieve similar overall accuracy
(49.7\% and 49.6\%) despite differing in architecture and serving
configuration. Gemma~4 scores 44.8\% (31B), 44.3\% (26B-A4B,
3.8B active), and 38.8\% (12B), although these configurations also
differ in architecture and precision. Qwen3.6-27B leads on L1--L4
and 34 sub-benchmarks, with the fewest unparsed replies among Qwen
and Gemma models. Its 5.9-point lead over Qwen3.8-27B decreases to
approximately 3.4 points when considering parsed replies alone,
but this comparison is affected by response selection.
Keye-VL-1.5-8B receives about 17.9k prompt tokens per question and
scores 41.3\%, compared with 1.5k tokens and 46.8\% for Molmo2-8B.
Differences in architectures, frame sampling, and serving configurations
prevent attributing performance differences to video-token budgets.

\begin{table}[h]

\caption{\textbf{Accuracy by ladder rung (\%, strict).}
Left: raw accuracy. Right: baseline-adjusted accuracy, computed
within each sub-benchmark and rung. L1 observation, L2 relations,
L3 measurement, L4 multi-step integration, L5 counterfactual.}
\label{tab:ladder}
\centering
\scriptsize
\setlength{\tabcolsep}{2.2pt}
\begin{tabular}{@{}lrrrrr|rrrrr@{}}
\toprule
& \multicolumn{5}{c|}{\textbf{Raw}} & \multicolumn{5}{c}{\textbf{Adjusted}}\\
\textbf{Model} & \textbf{L1} & \textbf{L2} & \textbf{L3} & \textbf{L4} & \textbf{L5}
& \textbf{L1} & \textbf{L2} & \textbf{L3} & \textbf{L4} & \textbf{L5}\\
\midrule
\rowcolor{rowgrey}
Qwen3.6-27B & \textbf{73.3} & \textbf{68.1} & \textbf{46.0} & \textbf{49.8} & 39.2 & \textbf{62.8} & \textbf{55.0} & \textbf{25.4} & \textbf{30.3} & 15.4\\
Qwen3.8-27B & 68.5 & 60.8 & 42.7 & 42.9 & 31.5 & 56.3 & 44.6 & 21.0 & 20.8 & 4.8\\
\rowcolor{rowgrey}
Qwen3.8-Flash-Next & 69.5 & 60.9 & 40.8 & 43.5 & 32.2 & 57.5 & 44.7 & 18.2 & 21.6 & 5.5\\
Molmo2-8B & 60.0 & 53.1 & 41.6 & 41.1 & 37.6 & 44.4 & 33.9 & 19.3 & 18.3 & 13.2\\
\rowcolor{rowgrey}
Nemotron-3.5-Super-VL & 57.3 & 51.0 & 39.2 & 41.0 & \textbf{39.7} & 40.6 & 30.9 & 16.0 & 18.1 & \textbf{16.0}\\
Gemma-4-31B & 63.1 & 57.1 & 38.1 & 36.1 & 27.3 & 48.8 & 39.3 & 14.4 & 11.3 & -1.2\\
\rowcolor{rowgrey}
Gemma-4-26B-A4B & 58.6 & 52.2 & 38.9 & 38.1 & 32.4 & 42.5 & 32.5 & 15.6 & 14.1 & 6.0\\
Keye-VL-1.5-8B & 56.4 & 46.7 & 35.7 & 35.6 & 32.0 & 39.3 & 24.8 & 11.1 & 10.6 & 5.3\\
\rowcolor{rowgrey}
Gemma-4-12B & 51.6 & 44.3 & 34.0 & 33.8 & 29.7 & 32.7 & 21.5 & 8.9 & 8.2 & 2.2\\
InternVideo2.5-8B & 44.6 & 42.0 & 32.9 & 32.3 & 32.0 & 23.0 & 18.3 & 7.3 & 6.2 & 5.5\\
\rowcolor{rowgrey}
Reka-Edge-2603 & 34.1 & 34.9 & 27.5 & 29.7 & 29.5 & 8.4 & 8.5 & -0.0 & 2.5 & 1.9\\
\midrule
Constant answer & 28.1 & 28.8 & 27.6 & 27.9 & 28.1 & 0.0 & 0.0 & 0.0 & 0.0 & 0.0\\
 
\bottomrule
\end{tabular}
\end{table}

\begin{table}[h]
\caption{\textbf{Accuracy by reasoning family} (\%, strict). Each family is
the mean of its sub-benchmarks, and the last column is the mean over
families.}
\label{tab:families_full}
\centering
\scriptsize
\setlength{\tabcolsep}{1.9pt}
\begin{tabular}{@{}lrrrrrrrr@{}}
\toprule
\textbf{Model} & 
\textbf{Mech.} & \textbf{Fluid} & \textbf{Spat.} & \textbf{Occl.} & \textbf{Temp.} & \textbf{Agent} & \textbf{Rules}
  & \textbf{Mean}\\
\midrule
\rowcolor{rowgrey}
Qwen3.6-27B & \textbf{54.9} & \textbf{63.3} & \textbf{47.3} & \textbf{58.2} & \textbf{51.9} & \textbf{56.8} & \textbf{53.5} & 55.1\\
Qwen3.8-27B & 49.3 & 55.3 & 43.8 & 53.8 & 46.3 & 48.8 & 47.9 & 49.3\\
\rowcolor{rowgrey}
Qwen3.8-Flash-Next & 50.6 & 59.3 & 41.4 & 51.7 & 43.9 & 51.5 & 44.4 & 49.0\\
Molmo2-8B & 46.8 & 47.8 & 41.3 & 49.7 & 45.5 & 46.9 & 47.8 & 46.5\\
\rowcolor{rowgrey}
Nemotron-3.5-Super-VL & 47.1 & 46.8 & 41.5 & 49.0 & 42.8 & 46.0 & 42.8 & 45.2\\
Gemma-4-31B & 45.8 & 46.4 & 43.4 & 51.8 & 38.2 & 49.3 & 36.4 & 44.5\\
\rowcolor{rowgrey}
Gemma-4-26B-A4B & 45.1 & 45.9 & 40.0 & 47.6 & 41.0 & 46.8 & 41.4 & 44.0\\
Keye-VL-1.5-8B & 42.4 & 41.8 & 38.0 & 43.9 & 38.8 & 42.7 & 39.3 & 41.0\\
\rowcolor{rowgrey}
Gemma-4-12B & 40.1 & 36.2 & 39.8 & 44.5 & 38.6 & 36.4 & 34.6 & 38.6\\
InternVideo2.5-8B & 36.7 & 36.5 & 35.1 & 39.8 & 37.0 & 37.5 & 34.8 & 36.8\\
\rowcolor{rowgrey}
Reka-Edge-2603 & 32.0 & 29.6 & 30.6 & 32.7 & 31.0 & 30.5 & 31.1 & 31.1\\
\midrule
Constant answer & 28.5 & 27.3 & 27.3 & 29.1 & 28.6 & 27.3 & 28.1 & 28.0\\
 
\bottomrule
\end{tabular}
\end{table}

\begin{table}[h]
\caption{\textbf{Motion-dependent versus single-frame questions}
(\%, strict). \emph{Motion}: the 81{,}108 questions flagged
as requiring motion. \emph{Frame}: the other 18{,}892.}
\label{tab:motion}
\centering
\footnotesize
\begin{tabular}{@{}lrrr@{}}
\toprule
\textbf{Model} & \textbf{Motion} & \textbf{Frame} & \textbf{Gap}\\
\midrule
\rowcolor{rowgrey}
Qwen3.6-27B & 52.7 & 68.1 & 15.4\\
Qwen3.8-27B & 46.7 & 62.5 & 15.8\\
\rowcolor{rowgrey}
Qwen3.8-Flash-Next & 46.0 & 65.1 & 19.1\\
Molmo2-8B & 44.7 & 55.9 & 11.2\\
\rowcolor{rowgrey}
Nemotron-3.5-Super-VL & 43.1 & 56.0 & 12.9\\
Gemma-4-31B & 42.2 & 56.3 & 14.1\\
\rowcolor{rowgrey}
Gemma-4-26B-A4B & 42.0 & 54.4 & 12.4\\
Keye-VL-1.5-8B & 38.6 & 52.7 & 14.1\\
\rowcolor{rowgrey}
Gemma-4-12B & 36.4 & 49.0 & 12.6\\
InternVideo2.5-8B & 35.4 & 43.1 & 7.7\\
\rowcolor{rowgrey}
Reka-Edge-2603 & 30.5 & 34.0 & 3.5\\
 
\bottomrule
\end{tabular}
\end{table}

\begin{table*}[h]
\caption{\textbf{Evaluation conditions per model.} \emph{Access}: hosted
service (Aleph, NRP) or our own vLLM server. \emph{Video input}: frames the
model receives; on the hosted services, frame sampling is done by the server
and is not reported. \emph{Thinking off}: how reasoning mode was disabled, by a request flag or
a prompt suffix (--: the model has no such mode). \emph{Unparsed}: replies without a leading
option letter (\%, scored as wrong). All models use temperature~0, at most 16
output tokens, and the same prompt text.}
\label{tab:conditions}
\centering
\scriptsize
\setlength{\tabcolsep}{2pt}
\begin{tabular}{@{}llllllr@{}}
\toprule
\textbf{Model} & \textbf{Access} & \textbf{Video input} & \textbf{Precision} & \textbf{Thinking off} & \textbf{System prompt} & \textbf{Unparsed}\\
\midrule
\rowcolor{rowgrey}
Qwen3.6-27B & Aleph & server-sampled & BF16 & request flag & none & 1.2\\
Qwen3.8-27B & Aleph & server-sampled & FP8 & request flag & none & 5.8\\
\rowcolor{rowgrey}
Gemma-4-26B-A4B & Aleph & server-sampled & FP8 & request flag & none & 2.3\\
Molmo2-8B & Aleph & server-sampled & not reported & request flag & none & 0.0\\
\rowcolor{rowgrey}
Qwen3.8-Flash-Next & NRP & server-sampled & FP8 & request flag & none & 5.1\\
Gemma-4-31B & NRP & server-sampled & W4A16 & request flag & none & 15.7\\
\rowcolor{rowgrey}
Gemma-4-12B & NRP & server-sampled & W4A16 & request flag & none & 7.1\\
Nemotron-3.5-Super-VL & own, vLLM 0.32, 2 GPUs & 2\,fps, $\le$256 frames, 50\% token pruning & NVFP4 & request flag & none & 0.0\\
\rowcolor{rowgrey}
Keye-VL-1.5-8B & own, vLLM 0.32 & 2\,fps, $\le$256 frames & BF16 & prompt suffix & none & 0.0\\
InternVideo2.5-8B & own, vLLM 0.32 + plugin & 32 frames, HiCo merging & BF16 & -- & checkpoint default & 0.0\\
\rowcolor{rowgrey}
Reka-Edge-2603 & own, vLLM 0.17.1 + plugin & 16 frames & BF16 & -- & none & 0.0\\
\bottomrule
\end{tabular}
\end{table*}

\begin{table}[h]
\caption{\textbf{Ranking under different scoring choices} (\%, rank in
parentheses). \emph{Parsed only}: accuracy on replies with a leading letter;
models are scored on different subsets of questions.}
\label{tab:rank_robust}
\centering
\scriptsize
\setlength{\tabcolsep}{3pt}
\begin{tabular}{@{}lrrr@{}}
\toprule
\textbf{Model} & \textbf{Strict} & \textbf{Lenient} & \textbf{Parsed only}\\
\midrule
\rowcolor{rowgrey}
Qwen3.6-27B & 55.6 (1) & 55.6 (1) & 56.2 (1)\\
Qwen3.8-27B & 49.7 (2) & 49.7 (2) & 52.8 (3)\\
\rowcolor{rowgrey}
Qwen3.8-Flash-Next & 49.6 (3) & 49.6 (3) & 52.3 (4)\\
Molmo2-8B & 46.8 (4) & 46.8 (4) & 46.8 (5)\\
\rowcolor{rowgrey}
Nemotron-3.5-Super-VL & 45.6 (5) & 45.6 (5) & 45.6 (6)\\
Gemma-4-31B & 44.8 (6) & 45.1 (6) & 53.2 (2)\\
\rowcolor{rowgrey}
Gemma-4-26B-A4B & 44.3 (7) & 44.3 (7) & 45.3 (7)\\
Keye-VL-1.5-8B & 41.3 (8) & 41.3 (8) & 41.3 (9)\\
\rowcolor{rowgrey}
Gemma-4-12B & 38.8 (9) & 39.7 (9) & 41.8 (8)\\
InternVideo2.5-8B & 36.9 (10) & 36.9 (10) & 36.9 (10)\\
\rowcolor{rowgrey}
Reka-Edge-2603 & 31.2 (11) & 31.2 (11) & 31.2 (11)\\
\midrule
Kendall's $\tau$ vs.\ strict & -- & 1.00 & 0.82\\
 
\bottomrule
\end{tabular}
\end{table}

\begin{table}[h]
\caption{\textbf{Motion gap, raw and at matched difficulty} (percentage
points; accuracy on unflagged minus flagged questions). \emph{Matched}: the
gap inside each (sub-benchmark, rung) cell that contains both kinds of
question (84 cells), weighted by the smaller count in the cell.}
\label{tab:motion_strat}
\centering
\scriptsize
\begin{tabular}{@{}lrr@{}}
\toprule
\textbf{Model} & \textbf{Raw} & \textbf{Matched}\\
\midrule
\rowcolor{rowgrey}
Qwen3.6-27B & 15.4 & 6.2\\
Qwen3.8-27B & 15.8 & 6.4\\
\rowcolor{rowgrey}
Qwen3.8-Flash-Next & 19.1 & 8.0\\
Molmo2-8B & 11.2 & 4.1\\
\rowcolor{rowgrey}
Nemotron-3.5-Super-VL & 12.9 & 7.9\\
Gemma-4-31B & 14.1 & 3.8\\
\rowcolor{rowgrey}
Gemma-4-26B-A4B & 12.4 & 4.8\\
Keye-VL-1.5-8B & 14.1 & 5.8\\
\rowcolor{rowgrey}
Gemma-4-12B & 12.6 & 9.1\\
InternVideo2.5-8B & 7.7 & 3.5\\
\rowcolor{rowgrey}
Reka-Edge-2603 & 3.5 & 1.5\\
 
\bottomrule
\end{tabular}
\end{table}

\begin{table}[h]
\caption{\textbf{Injected-violation questions} (\%, 301 questions, 97 from
violation clips). F1: detecting that a rule was broken. Loc.: picking the
exact culprit on violation clips. Clean: picking ``nothing went wrong'' on
clips without a violation. With 20--52 violation questions per
sub-benchmark, these numbers carry wide intervals.}
\label{tab:violation}
\centering
\scriptsize
\setlength{\tabcolsep}{1.7pt}
\begin{tabular}{@{}lrr|rr|rr|rrr@{}}
\toprule
& \multicolumn{2}{c|}{\textbf{Chess}} & \multicolumn{2}{c|}{\textbf{Symm.}} & \multicolumn{2}{c|}{\textbf{Flood}} & \multicolumn{3}{c}{\textbf{All}}\\
\textbf{Model} & \textbf{F1} & \textbf{Loc.} & \textbf{F1} & \textbf{Loc.} & \textbf{F1} & \textbf{Loc.} & \textbf{F1} & \textbf{Loc.} & \textbf{Clean}\\
\midrule
\rowcolor{rowgrey}
Qwen3.6-27B & 0.0 & 0.0 & 45.2 & 15.0 & 30.2 & 3.8 & 28.0 & 5.2 & 87.3\\
Qwen3.8-27B & 38.2 & 16.0 & 33.3 & 20.0 & 37.6 & 13.5 & 36.7 & 15.5 & 31.4\\
\rowcolor{rowgrey}
Qwen3.8-Flash-Next & 42.9 & 52.0 & 39.4 & 30.0 & 51.8 & 38.5 & 46.6 & 40.2 & 14.2\\
Molmo2-8B & 7.7 & 0.0 & 9.5 & 5.0 & 30.4 & 19.2 & 23.0 & 11.3 & 87.3\\
\rowcolor{rowgrey}
Nemotron-3.5-Super-VL & 33.3 & 16.0 & 35.9 & 0.0 & 47.8 & 23.1 & 42.8 & 16.5 & 64.7\\
Gemma-4-31B & 19.5 & 4.0 & 35.1 & 25.0 & 49.7 & 28.8 & 41.4 & 21.6 & 38.2\\
\rowcolor{rowgrey}
Gemma-4-26B-A4B & 38.0 & 32.0 & 39.5 & 30.0 & 54.3 & 38.5 & 47.2 & 35.1 & 18.1\\
Keye-VL-1.5-8B & 0.0 & 0.0 & 8.7 & 0.0 & 38.5 & 13.5 & 27.6 & 7.2 & 83.3\\
\rowcolor{rowgrey}
Gemma-4-12B & 12.9 & 0.0 & 39.1 & 15.0 & 47.8 & 23.1 & 40.8 & 15.5 & 64.2\\
InternVideo2.5-8B & 44.4 & 32.0 & 9.5 & 5.0 & 43.5 & 23.1 & 40.9 & 21.6 & 55.9\\
\rowcolor{rowgrey}
Reka-Edge-2603 & 0.0 & 0.0 & 0.0 & 0.0 & 49.4 & 26.9 & 38.9 & 14.4 & 60.3\\
 
\bottomrule
\end{tabular}
\end{table}

\begin{table*}[p]
\caption{\textbf{Accuracy per sub-benchmark} (\%, strict, 2{,}500 questions
per model and sub-benchmark). \emph{Base}: constant-answer baseline.
Bold: best model.}
\label{tab:percat}
\centering
\scriptsize
\setlength{\tabcolsep}{4.5pt}
\begin{tabular}{@{}lrrrrrrrrrrrrr@{}}
\toprule
\textbf{Sub-benchmark} & \textbf{Base} & 
\rotatebox{90}{Qwen3.6-27B} & \rotatebox{90}{Qwen3.8-27B} & \rotatebox{90}{Qwen3.8-Flash-Next} & \rotatebox{90}{Molmo2-8B} & \rotatebox{90}{Nemotron-3.5-Super-VL} & \rotatebox{90}{Gemma-4-31B} & \rotatebox{90}{Gemma-4-26B-A4B} & \rotatebox{90}{Keye-VL-1.5-8B} & \rotatebox{90}{Gemma-4-12B} & \rotatebox{90}{InternVideo2.5-8B} & \rotatebox{90}{Reka-Edge-2603}
  & \textbf{Mean}\\
\midrule
\multicolumn{14}{@{}l}{\emph{Mechanics \& motion}}\\
\rowcolor{rowgrey}
bouncing ball restitution & 26.6 & \textbf{53.2} & 49.6 & 50.0 & 50.0 & 51.4 & 45.9 & 44.6 & 42.6 & 38.6 & 36.7 & 32.2 & 45.0\\
shape collision & 27.9 & \textbf{46.1} & 41.4 & 41.0 & 45.8 & 42.7 & 36.4 & 39.8 & 42.6 & 34.3 & 35.5 & 27.2 & 39.4\\
\rowcolor{rowgrey}
gear train & 27.8 & \textbf{58.8} & 58.3 & 54.7 & 42.3 & 48.8 & 56.4 & 51.7 & 44.6 & 45.1 & 33.0 & 30.9 & 47.7\\
simple harmonic motion & 28.0 & 42.4 & 38.2 & 40.4 & \textbf{46.4} & 45.2 & 30.1 & 38.8 & 38.3 & 33.4 & 36.0 & 29.4 & 38.1\\
\rowcolor{rowgrey}
spring chain waves & 26.9 & \textbf{53.1} & 45.6 & 46.8 & 47.1 & 46.1 & 41.6 & 44.3 & 37.0 & 39.6 & 31.8 & 29.5 & 42.0\\
balance scale weighing & 26.8 & \textbf{70.0} & 63.2 & 63.8 & 54.1 & 54.3 & 59.4 & 53.0 & 54.8 & 46.8 & 45.7 & 33.4 & 54.4\\
\rowcolor{rowgrey}
rolling races & 26.3 & \textbf{50.6} & 42.6 & 45.8 & 37.4 & 38.4 & 41.0 & 36.6 & 31.8 & 36.2 & 27.7 & 31.0 & 38.1\\
inertia race & 30.6 & \textbf{55.2} & 41.4 & 53.9 & 45.0 & 49.7 & 44.6 & 46.0 & 41.5 & 43.2 & 37.6 & 36.8 & 45.0\\
\rowcolor{rowgrey}
stacking collapse & 35.4 & \textbf{64.8} & 63.5 & 58.8 & 52.9 & 47.8 & 56.4 & 50.6 & 48.4 & 43.8 & 46.6 & 37.8 & 51.9\\
\midrule
\multicolumn{14}{@{}l}{\emph{Fluids \& optics}}\\
\rowcolor{rowgrey}
buoyancy tank & 26.7 & \textbf{64.4} & 60.0 & 63.9 & 45.9 & 53.7 & 54.2 & 49.8 & 46.8 & 44.8 & 38.8 & 32.7 & 50.4\\
vessels and clocks & 26.7 & \textbf{68.6} & 57.2 & 61.2 & 47.6 & 43.9 & 50.0 & 47.3 & 42.8 & 34.5 & 39.1 & 35.2 & 48.0\\
\rowcolor{rowgrey}
glasses fill & 27.5 & \textbf{82.0} & 75.7 & 78.9 & 55.4 & 46.1 & 63.0 & 52.8 & 48.0 & 36.5 & 30.8 & 23.1 & 53.8\\
well flood trace & 27.1 & \textbf{60.2} & 54.0 & 51.6 & 39.8 & 47.2 & 47.1 & 45.0 & 35.8 & 27.1 & 34.7 & 25.9 & 42.6\\
\rowcolor{rowgrey}
mirror beam reflection & 28.0 & \textbf{52.5} & 39.7 & 50.8 & 52.3 & 45.7 & 30.2 & 37.6 & 41.3 & 34.8 & 35.1 & 30.4 & 40.9\\
hidden mirrors & 27.6 & \textbf{52.4} & 45.2 & 49.5 & 45.7 & 44.1 & 34.1 & 42.8 & 35.8 & 39.6 & 40.4 & 30.5 & 41.8\\
\midrule
\multicolumn{14}{@{}l}{\emph{Spatial \& 3D geometry}}\\
\rowcolor{rowgrey}
spinning polyhedra & 27.0 & 43.0 & 42.9 & 37.8 & 41.8 & 42.2 & \textbf{48.0} & 40.8 & 38.0 & 38.6 & 34.7 & 28.3 & 39.7\\
turntable voxels & 27.8 & \textbf{47.7} & 47.1 & 39.1 & 39.4 & 43.8 & 40.2 & 44.2 & 35.8 & 43.1 & 36.6 & 33.6 & 41.0\\
\rowcolor{rowgrey}
moving shadows & 27.6 & \textbf{40.8} & 35.2 & 38.4 & 31.6 & 31.4 & 34.8 & 32.5 & 33.3 & 31.1 & 28.7 & 24.3 & 32.9\\
symmetry pattern video & 26.6 & \textbf{57.5} & 50.0 & 50.2 & 52.2 & 48.6 & 50.4 & 42.6 & 44.8 & 46.4 & 40.5 & 36.0 & 47.2\\
\midrule
\multicolumn{14}{@{}l}{\emph{Occlusion \& permanence}}\\
\rowcolor{rowgrey}
occluded pattern counting & 28.6 & \textbf{56.5} & 53.0 & 46.5 & 45.1 & 47.2 & 53.0 & 48.2 & 40.2 & 42.3 & 38.6 & 36.0 & 46.0\\
translucent occlusion & 27.4 & \textbf{69.5} & 61.2 & 59.9 & 52.6 & 49.4 & 64.7 & 55.6 & 47.6 & 49.2 & 43.7 & 35.7 & 53.5\\
\rowcolor{rowgrey}
overlapping shapes & 27.6 & \textbf{68.8} & 64.8 & 68.4 & 57.0 & 60.8 & 57.3 & 53.4 & 52.1 & 52.4 & 44.7 & 28.2 & 55.3\\
shell game tracking & 36.4 & \textbf{49.6} & 48.5 & 40.5 & 42.8 & 41.6 & 42.2 & 40.8 & 39.3 & 42.2 & 37.8 & 36.8 & 42.0\\
\rowcolor{rowgrey}
object tracking mot & 27.0 & \textbf{62.8} & 60.4 & 57.5 & 58.2 & 56.5 & 62.3 & 54.4 & 50.3 & 47.6 & 43.7 & 31.0 & 53.2\\
rainbow bagel & 27.9 & 41.8 & 35.0 & 37.4 & \textbf{42.7} & 38.6 & 31.0 & 33.0 & 33.9 & 33.1 & 30.2 & 28.4 & 35.0\\
\midrule
\multicolumn{14}{@{}l}{\emph{Temporal events}}\\
\rowcolor{rowgrey}
temporal flash counting & 27.7 & 48.9 & 44.3 & 40.2 & \textbf{51.5} & 41.8 & 39.9 & 40.2 & 42.1 & 39.7 & 38.6 & 28.8 & 41.5\\
nested shapes & 27.6 & \textbf{50.8} & 46.7 & 47.2 & 41.9 & 40.9 & 44.3 & 40.9 & 42.2 & 41.6 & 36.3 & 30.4 & 42.1\\
\rowcolor{rowgrey}
blinking networks & 27.2 & \textbf{59.0} & 50.1 & 51.5 & 46.0 & 44.7 & 48.2 & 45.4 & 41.0 & 39.4 & 31.0 & 25.9 & 43.9\\
sandpile cascade & 32.8 & \textbf{49.8} & 43.2 & 38.9 & 45.7 & 41.4 & 24.8 & 38.9 & 33.2 & 37.4 & 40.3 & 35.7 & 39.0\\
\rowcolor{rowgrey}
cellular automata & 27.7 & \textbf{50.9} & 47.2 & 41.7 & 42.6 & 45.0 & 33.8 & 39.4 & 35.2 & 34.7 & 38.6 & 34.2 & 40.3\\
\midrule
\multicolumn{14}{@{}l}{\emph{Agents \& navigation}}\\
\rowcolor{rowgrey}
multi agent chase & 27.7 & \textbf{61.0} & 53.2 & 47.7 & 47.3 & 45.3 & 47.7 & 47.6 & 44.4 & 37.8 & 38.4 & 30.8 & 45.6\\
colored route puzzle & 27.7 & \textbf{59.6} & 54.4 & 58.0 & 49.1 & 50.9 & 55.8 & 51.8 & 43.0 & 43.9 & 37.4 & 26.1 & 48.2\\
\rowcolor{rowgrey}
traffic light queue & 27.2 & 63.7 & 57.6 & 59.8 & 54.2 & 52.0 & \textbf{64.4} & 56.0 & 47.8 & 47.5 & 40.9 & 34.9 & 52.6\\
swing time & 27.0 & \textbf{52.7} & 36.6 & 47.0 & 42.6 & 41.0 & 36.1 & 35.9 & 39.0 & 29.5 & 34.8 & 26.0 & 38.3\\
\rowcolor{rowgrey}
robot long jump & 26.9 & \textbf{47.0} & 42.0 & 45.0 & 41.5 & 40.8 & 42.7 & 42.7 & 39.4 & 23.0 & 36.3 & 34.7 & 39.5\\
\midrule
\multicolumn{14}{@{}l}{\emph{Rules \& algorithms}}\\
\rowcolor{rowgrey}
chess dance trace & 26.7 & \textbf{63.0} & 53.8 & 52.9 & 46.0 & 44.8 & 41.5 & 43.7 & 43.5 & 35.8 & 33.5 & 28.2 & 44.2\\
tangled braid & 28.4 & 52.2 & 50.4 & 46.1 & \textbf{53.6} & 48.0 & 40.3 & 40.8 & 40.5 & 38.3 & 34.7 & 31.8 & 43.4\\
\rowcolor{rowgrey}
tile alignment shuffle & 26.9 & \textbf{46.3} & 40.5 & 34.8 & 42.4 & 38.6 & 16.9 & 34.5 & 32.9 & 23.8 & 32.5 & 29.9 & 33.9\\
conveyor belt sort & 30.4 & \textbf{54.8} & 51.6 & 45.6 & 51.3 & 42.4 & 40.2 & 42.1 & 41.0 & 36.6 & 37.4 & 35.5 & 43.5\\
\rowcolor{rowgrey}
robot archery & 28.0 & \textbf{50.9} & 43.0 & 42.4 & 45.6 & 40.4 & 43.0 & 45.9 & 38.6 & 38.5 & 35.8 & 30.0 & 41.3\\
 
\bottomrule
\end{tabular}
\end{table*}

\section{Validation details}
\label{sec:validation}

\cref{tab:validation} summarizes four checks per sub-benchmark.

\paragraph{Release validation.} The release validator rebuilds every
sample from its seed, re-derives every answer with the sub-benchmark's answer
module, and checks the ladder, the simulation invariants, and every video
(all frames decode; resolution matches; duration within 0.2\,s of the plan;
mean absolute frame difference above 0.05). Over all $100{,}000$ samples and
$1{,}000{,}000$ questions it reports no error and zero mismatches. Two
reports predated a wording fix to their annotation files
(\emph{balance scale weighing}, \emph{glasses fill}) and were re-run on the
released files.

\paragraph{Pixel checks.} The 26 sub-benchmarks with a frame checker were
checked on their first 50 clips ($1{,}300$ clips, $239{,}677$ decoded frames).
Each checker locates the drawn objects, clock, and labels and compares them
with the scene state within a stated tolerance, typically 2--4\,px for
positions; 11 also re-derive the answers of the checked clips from the
decoded frames. The other 14 sub-benchmarks
rely on the generic video checks above.

\paragraph{Tamper test.} For each sub-benchmark we copied five clips, gave
three questions a wrong but self-consistent answer (stored letter, answer
text, and ground truth), and re-ran the validator: it flagged all 120
corrupted questions. Corrupting only the evaluation file
raised no flag, because the validator reads only
the annotation file. A separate comparison therefore checked the two files
directly: for all $1{,}000{,}000$ questions they agree on IDs, answer letters,
answer texts, options, and question text.

\paragraph{Independence of the answer path.} Each answer module re-derives
the answers from the drawn scene, usually by reading the renderer's
per-frame geometry, so the pixel checks above tie both to the video. An
import audit shows that 26 answer modules use no simulation or question code
from the generator, six use only text, clock, or constant helpers, four reuse
geometry primitives (\emph{moving shadows}, \emph{nested shapes},
\emph{overlapping shapes}, \emph{rolling races}), and four call generator
state or simulation functions (\emph{multi-agent chase}, \emph{occluded
pattern counting}, \emph{translucent occlusion}, \emph{spinning polyhedra}).
For these last four, agreement between generator and validator is weaker
evidence, since shared code can repeat a mistake.

\begin{table}[t]
\caption{\textbf{Validation per sub-benchmark.} \emph{Mism.}: answers that
differ from the re-derived ones over all $25{,}000$ questions. \emph{Clips},
\emph{Frames}, \emph{Fail.}: pixel check on the first 50 clips; each sub-benchmark's
failures fall on a single clip, and those inspected (\emph{chess dance
trace}, \emph{rolling races}, \emph{shape collision}) are limits of the
checker, not of the data. \emph{Tamper}: corrupted answers flagged. \emph{Shared}:
generator code used by the answer module beyond the renderer (text: text,
clock, or constant helpers; geom.: geometry primitives; state: simulation
or state functions). $^\dagger$: answer module does not use the renderer.}
\label{tab:validation}
\centering
\scriptsize
\setlength{\tabcolsep}{3pt}
\begin{tabular}{@{}lrrrrcl@{}}
\toprule
\textbf{Sub-benchmark} & \textbf{Mism.} & \textbf{Clips} & \textbf{Frames} & \textbf{Fail.} & \textbf{Tamper} & \textbf{Shared}\\
\midrule
\rowcolor{rowgrey}
balance scale weighing & 0 & -- & -- & -- & 3/3 & --\\
blinking networks$^\dagger$ & 0 & -- & -- & -- & 3/3 & --\\
\rowcolor{rowgrey}
bouncing ball restitution & 0 & -- & -- & -- & 3/3 & --\\
buoyancy tank & 0 & -- & -- & -- & 3/3 & --\\
\rowcolor{rowgrey}
cellular automata & 0 & 50 & 9{,}360 & 0 & 3/3 & text\\
chess dance trace & 0 & 50 & 9{,}029 & 2 & 3/3 & --\\
\rowcolor{rowgrey}
colored route puzzle & 0 & 50 & 8{,}706 & 1 & 3/3 & --\\
conveyor belt sort & 0 & 50 & 10{,}617 & 0 & 3/3 & --\\
\rowcolor{rowgrey}
gear train & 0 & -- & -- & -- & 3/3 & text\\
glasses fill & 0 & 50 & 8{,}715 & 0 & 3/3 & --\\
\rowcolor{rowgrey}
hidden mirrors & 0 & 50 & 8{,}572 & 0 & 3/3 & --\\
inertia race & 0 & -- & -- & -- & 3/3 & --\\
\rowcolor{rowgrey}
mirror beam reflection & 0 & 50 & 6{,}289 & 0 & 3/3 & --\\
moving shadows & 0 & 50 & 10{,}261 & 0 & 3/3 & geom.\\
\rowcolor{rowgrey}
multi agent chase & 0 & 50 & 10{,}102 & 0 & 3/3 & \textbf{state}\\
nested shapes & 0 & 50 & 9{,}922 & 0 & 3/3 & geom.\\
\rowcolor{rowgrey}
object tracking mot & 0 & 50 & 9{,}219 & 0 & 3/3 & --\\
occluded pattern counting & 0 & 50 & 8{,}090 & 0 & 3/3 & \textbf{state}\\
\rowcolor{rowgrey}
overlapping shapes$^\dagger$ & 0 & -- & -- & -- & 3/3 & geom.\\
rainbow bagel & 0 & -- & -- & -- & 3/3 & text\\
\rowcolor{rowgrey}
robot archery & 0 & 50 & 7{,}500 & 0 & 3/3 & --\\
robot long jump & 0 & -- & -- & -- & 3/3 & --\\
\rowcolor{rowgrey}
rolling races & 0 & 50 & 9{,}344 & 1 & 3/3 & geom.\\
sandpile cascade & 0 & 50 & 10{,}256 & 0 & 3/3 & text\\
\rowcolor{rowgrey}
shape collision & 0 & 50 & 10{,}771 & 18 & 3/3 & --\\
shell game tracking & 0 & 50 & 10{,}056 & 0 & 3/3 & text\\
\rowcolor{rowgrey}
simple harmonic motion & 0 & 50 & 9{,}664 & 0 & 3/3 & --\\
spinning polyhedra & 0 & 50 & 8{,}909 & 0 & 3/3 & \textbf{state}\\
\rowcolor{rowgrey}
spring chain waves & 0 & 50 & 10{,}012 & 0 & 3/3 & --\\
stacking collapse & 0 & 50 & 7{,}155 & 0 & 3/3 & --\\
\rowcolor{rowgrey}
swing time & 0 & -- & -- & -- & 3/3 & --\\
symmetry pattern video & 0 & 50 & 6{,}508 & 0 & 3/3 & --\\
\rowcolor{rowgrey}
tangled braid & 0 & -- & -- & -- & 3/3 & --\\
temporal flash counting & 0 & 50 & 9{,}579 & 0 & 3/3 & text\\
\rowcolor{rowgrey}
tile alignment shuffle & 0 & -- & -- & -- & 3/3 & --\\
traffic light queue & 0 & 50 & 10{,}336 & 0 & 3/3 & --\\
\rowcolor{rowgrey}
translucent occlusion & 0 & 50 & 10{,}304 & 0 & 3/3 & \textbf{state}\\
turntable voxels$^\dagger$ & 0 & -- & -- & -- & 3/3 & --\\
\rowcolor{rowgrey}
vessels and clocks & 0 & 50 & 10{,}401 & 0 & 3/3 & --\\
well flood trace$^\dagger$ & 0 & -- & -- & -- & 3/3 & --\\
\midrule
All 40 & 0 & 1{,}300 & 239{,}677 & 22 & 120/120 & 8\\
 
\bottomrule
\end{tabular}
\end{table}

\section{Human baseline}
\label{sec:human}

Three annotators answered $7{,}720$ questions on 772 clips, all drawn
from the $100{,}000$-question model evaluation, so humans and models are
scored on the same questions. The study covers 15 sub-benchmarks, two or three
per family, with 50 clips and 500 questions each; 22 further clips from
other sub-benchmarks count only toward the overall and per-rung numbers.
Each clip was answered by one annotator, who watched the video and answered
its ten questions; the median time per clip was 8.9\,s (10th--90th
percentile: 3.0--34.4\,s). Answers are scored strictly, as for the models.

Humans answer 91.2\% of all $7{,}720$ questions, against 61.1\% for the best
model, Qwen3.6-27B, on the same questions, and 91.5\% averaged over the 15
sub-benchmarks (\cref{tab:human}). They stay above 84\% on every rung, while
the best model on each rung drops from 81.7\% at L1 to 45.0\% at L5 (\cref{tab:human_rung}).
Humans score above the best model on every sub-benchmark. Their lowest
scores are on \emph{inertia race} (68.6\%) and \emph{symmetry pattern
video} (78.2\%); every other sub-benchmark is at least 84\%. The three
annotators score 88.1\%, 88.8\%, and 97.0\%. The gap between humans and
models is therefore not caused by unanswerable questions.

\begin{table}[t]
\caption{\textbf{Human baseline} (\%, strict; 50 clips and 500
questions per sub-benchmark, one annotator per question). The
constant-answer baseline is 27--31\% on every row. \emph{Models}: mean of the eleven
models; \emph{Best}: best model on that sub-benchmark; both on the same
questions as the humans.}
\label{tab:human}
\centering
\scriptsize
\setlength{\tabcolsep}{3.5pt}
\begin{tabular}{@{}llrrr@{}}
\toprule
& \textbf{Sub-benchmark} & \textbf{Models} & \textbf{Best} & \textbf{Human}\\
\midrule
Mech. & bouncing ball restitution & 46.0 & 57.0 & \textbf{98.6}\\
 & balance scale weighing & 55.2 & 70.6 & \textbf{100.0}\\
 & inertia race & 46.6 & 55.4 & \textbf{68.6}\\
\midrule
Fluid & buoyancy tank & 50.8 & 64.4 & \textbf{99.6}\\
 & glasses fill & 52.9 & 83.2 & \textbf{97.2}\\
\midrule
Spat. & turntable voxels & 41.3 & 52.2 & \textbf{84.0}\\
 & symmetry pattern video & 44.4 & 54.2 & \textbf{78.2}\\
\midrule
Occl. & overlapping shapes & 55.5 & 69.0 & \textbf{94.6}\\
 & object tracking mot & 52.2 & 62.8 & \textbf{84.4}\\
\midrule
Temp. & nested shapes & 40.6 & 48.4 & \textbf{97.2}\\
 & blinking networks & 42.5 & 59.2 & \textbf{97.8}\\
\midrule
Agent & colored route puzzle & 49.2 & 62.2 & \textbf{90.6}\\
 & traffic light queue & 52.5 & 66.6 & \textbf{94.4}\\
\midrule
Rules & chess dance trace & 44.0 & 62.2 & \textbf{90.6}\\
 & conveyor belt sort & 44.3 & 57.2 & \textbf{97.4}\\
\midrule
 & \textbf{All 15} & 47.9 & 61.6 & \textbf{91.5}\\
 
\bottomrule
\end{tabular}
\end{table}

\begin{table}[t]
\caption{\textbf{Human baseline by ladder rung} (\%, strict; all
$7{,}720$ questions; the constant-answer baseline is 27--28\% on every
rung). \emph{Best model}: the best of the eleven models on
each rung, on the same questions.}
\label{tab:human_rung}
\centering
\scriptsize
\begin{tabular}{@{}lrrr@{}}
\toprule
\textbf{Rung} & \textbf{Questions} & \textbf{Best model} & \textbf{Human}\\
\midrule
L1 & 1{,}373 & 81.7 & \textbf{95.3}\\
L2 & 1{,}849 & 74.8 & \textbf{94.1}\\
L3 & 1{,}868 & 49.7 & \textbf{90.5}\\
L4 & 1{,}382 & 54.6 & \textbf{90.4}\\
L5 & 1{,}248 & 45.0 & \textbf{84.2}\\
 
\bottomrule
\end{tabular}
\end{table}

\end{document}